\documentclass[acmtog,screen]{acmart}
\usepackage{color}
\usepackage{colortbl}
\usepackage{multirow}

\definecolor{a}{rgb}{1,0.710,0.639}
\definecolor{b}{rgb}{1,0.863,0.784}
\definecolor{c}{rgb}{1,1,0.784}

\newcommand{\name}{LUMEN}

\AtBeginDocument{%
  }

\usepackage{enumitem}

\copyrightyear{2026}
\acmYear{2026}
\setcopyright{cc}
\setcctype{by}
\acmConference[SA Conference Papers '26]{SIGGRAPH Asia 2026 Conference Papers}{December 01--04, 2026}{Kuala Lumpur, Malaysia}
\acmBooktitle{SIGGRAPH Asia 2026 Conference Papers (SA Conference Papers '26), December 01--04, 2026, Kuala Lumpur, Malaysia}
\acmDOI{10.1145/3829340.3842306}
\acmISBN{979-8-4007-2842-6/2026/12}
\begin{document}
\title{LUMEN: Geometry-Guided Diffusion Model for Spatially-Varying HDR Illumination Estimation}

\author{Hongze Chen}
\affiliation{%
  \institution{The Hong Kong University of Science and Technology}
  \country{Hong Kong}
}

\author{Fangyu Lin}
\affiliation{%
  \institution{The Hong Kong University of Science and Technology}
  \country{Hong Kong}
}

\author{Runze Zhang}
\affiliation{%
  \institution{LIGHTSPEED}
  \country{China}
}

\author{Lingting Zhu}
\affiliation{%
  \institution{LIGHTSPEED}
  \country{China}
}

\author{Keyang Luo}
\affiliation{%
  \institution{LIGHTSPEED}
  \country{China}
}

\author{Yingda Yin}
\affiliation{%
  \institution{LIGHTSPEED}
  \country{China}
}

\author{Zeyu Hu}
\affiliation{%
  \institution{LIGHTSPEED}
  \country{China}
}

\author{Weikai Chen}
\affiliation{%
  \institution{LIGHTSPEED}
  \country{China}
}

\author{Xin Wang}
\affiliation{%
  \institution{LIGHTSPEED}
  \country{China}
}

\author{Zehong Lin}
\affiliation{%
  \institution{Lingnan University}
  \country{Hong Kong}}

\author{Jun Zhang}
\authornote{Corresponding author.}
\affiliation{%
  \institution{The Hong Kong University of Science and Technology}
  \country{Hong Kong}}


\begin{abstract}
Estimating scene illumination from images is challenging as lighting is inherently spatially varying, depending on geometry, occlusion, and light sources. Existing methods incorporate geometric cues by deriving local environment maps or probe-based representations, but largely rely on regression or parametric models, resulting in over-smoothed lighting and limited realism.
We revisit spatially-varying illumination estimation from a generative perspective and introduce LUMEN, a geometry-conditioned diffusion framework for synthesizing high-dynamic-range (HDR) environment maps at arbitrary spatial locations from single or multi-view images.
Specifically, we reconstruct a coarse 3D scene from input images and render partial environment maps at arbitrary locations to encode visibility and spatial structure. These signals are fused with multi-view image features and fed into a diffusion transformer to generate full HDR environment maps.
To train the model, we build a synthetic dataset of 700 indoor scenes with dense spatial supervision, including multi-view RGB images, depth, and HDR illumination at multiple sampled locations. This dataset enables the model to learn spatially consistent lighting distributions across diverse scenes. Extensive experiments demonstrate that LUMEN significantly improves visual fidelity and spatial coherence compared to prior approaches. Project page: \href{https://stopaimme.github.io/lumen/}{https://stopaimme.github.io/lumen/}.
\end{abstract}


\begin{teaserfigure}
  \includegraphics[width=\textwidth]{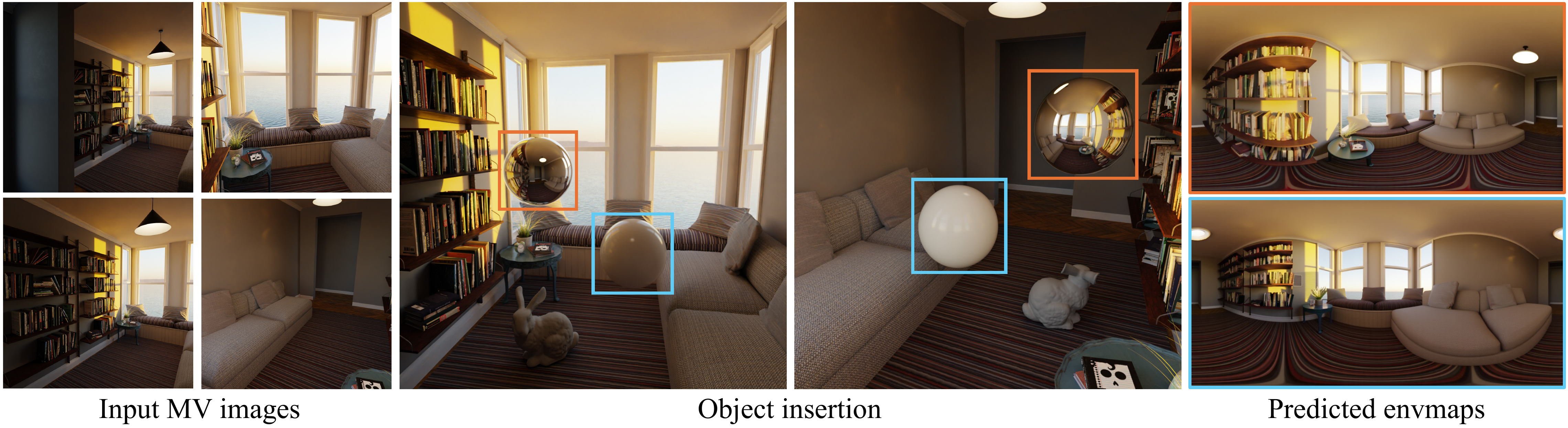}
  \vspace{-10pt}
  \caption{Given a set of multi-view images, LUMEN estimates spatially-varying illumination at any location with consistency and achieves seamless virtual object insertion results (the balls and the bunny are rendered using predicted HDRI maps at their positions).}
  \Description{teaser}
  \label{fig:teaser}
\end{teaserfigure}


\maketitle

\section{Introduction}

Estimating scene illumination from images is a fundamental problem in computer graphics and vision, 
with applications in augmented reality, relighting, and virtual object insertion. 
While many approaches model illumination as a single global environment map, 
real-world lighting is inherently spatially varying: 
illumination changes across space due to geometry, occlusion, and local light sources. 
Accurately recovering such spatially-varying illumination from limited visual observations 
remains a challenging and ill-posed problem.

Prior work has made progress by incorporating geometric reasoning into lighting estimation.
A common paradigm reconstructs scene structure and derives local lighting representations, such as per-pixel environment maps, probe-based lighting, or volumetric light fields at different spatial locations~\cite{garon2019fast,song2019neural,srinivasan2020lighthouse,gera2022casual,4dlighting,bolduc2025gaslight}.
These approaches introduce spatial awareness by explicitly modeling how illumination varies with position. 
However, they typically rely on regression models or low-dimensional parameterizations 
(e.g., spherical harmonics or Gaussian mixtures), which limits their ability to capture 
high-frequency lighting effects. 
As a result, the predicted illumination is often over-smoothed, lacks realism, 
and degrades significantly in regions that are weakly observed or not directly visible.



This limitation suggests that the core challenge lies not only in determining \emph{where} to estimate illumination, but also in \emph{how} to recover realistic lighting from incomplete local observations. In particular, local geometric cues, such as partial environment maps derived from reconstructed scenes, provide only a sparse and visibility-constrained view of the full lighting distribution. They capture visibility and occlusion but inevitably miss unseen regions and high-frequency lighting effects. Recovering complete high-dynamic-range (HDR) illumination from such partial observations requires strong priors over natural lighting, which are not well captured by regression-based methods. Thus, instead of direct regression, we propose reframing this task as a \emph{completion problem}, where the goal is to infer missing information about the full illumination distribution based on partial observations from a generative perspective.

To this end, we propose \emph{\name{}}, a geometry-conditioned diffusion framework that formulates illumination estimation as conditional generation. Instead of encoding spatial variation implicitly or conditioning on raw 3D coordinates, \name{} estimates illumination at a query location using a location-dependent partial environment map. Given the input views, it first recovers coarse scene geometry and camera poses via an off-the-shelf multi-view reconstruction method~\cite{VGGT}. For each query location, the reconstructed geometry is projected into a panoramic representation centered at that location, yielding a \emph{partial environment map} that captures geometry-aware visibility and occlusion cues. This partial environment map serves as an \emph{explicit spatial conditioning signal}, so changing the query location directly changes the conditioning input and thus the generated illumination.

We implement this design by fine-tuning a pre-trained diffusion transformer (DiT).
Multi-view image features and the location-dependent partial environment map are jointly fused with the noisy latent through the global attention mechanisms of the DiT, enabling effective information sharing across views while enforcing spatially {consistent} illumination estimation.
As a result, \name{} generates complete HDRI maps that are consistent with both appearance cues and spatial constraints.
Training a geometry-conditioned generative model requires supervision that captures how illumination varies across space.
To bridge this gap, we build a synthetic dataset that contains approximately \emph{700 indoor scenes} with diverse layouts, textures, and lighting, generated using a modified version of Infinigen Indoors~\cite{Raistrick_2024_CVPR}. For each scene, we render multi-view RGB images with depth maps and HDRI maps at different positions. The camera trajectory and light-probe placement follow predefined rules to ensure physically plausible configurations and broad scene coverage, providing sufficient contextual cues about appearance, lighting, and geometry.

Extensive experiments on both synthetic and real-world datasets demonstrate that \name{} 
produces more photorealistic and spatially consistent illumination than prior approaches. 
In particular, our method better captures high-frequency lighting variations, 
improves realism in unseen regions, and benefits from multi-view inputs to achieve 
stronger spatial coherence.
In summary, our main contributions are as follows:
\begin{itemize}[leftmargin = 15pt]
\item We reformulate spatially-varying illumination estimation as a geometry-conditioned 
generative problem, where lighting is recovered as completion from partial, visibility-constrained observations.
\item  We present \emph{\name{}}, a geometry-conditioned diffusion model that 
integrates partial environment maps as spatially grounded observations and leverages 
generative priors to infer missing regions and high-frequency illumination.
\item We construct a large-scale synthetic dataset with dense spatial 
illumination supervision, such as multi-view RGB images, depth maps, and spatially-varying environment illumination with camera parameters.
\item  We demonstrate consistent improvements in photorealism and spatial consistency 
over prior approaches.
\end{itemize}

\section{Related Work}

\paragraph{Lighting Estimation.} Estimating scene lighting has long been a fundamental challenge in computer vision and graphics, playing a critical role in applications such as virtual object insertion and relighting. Early methods mainly rely on physical light probes \cite{probe1} or specific scene cues, such as reflections in eyes \cite{eyes, Yi_2018_ECCV} or highlights from glossy objects \cite{chip}, to infer environmental illumination. To generalize beyond specific cues, learning-based methods have emerged to estimate lighting from casual visual captures. A prevalent paradigm is to directly regress HDRI maps from LDR images using neural networks trained on HDR datasets \cite{gardner2017learning, Hold-Geoffroy_2019_CVPR, legendre2019deeplight, Gardner_2019_ICCV, Zhang_2019_CVPR, LeGendre_SA_20, zhan2021emlight, Somanath2021HDREM}. While effective for global illumination estimation, these regression methods often struggle to provide high-frequency details in unobserved regions.

Recent advances in generative models have demonstrated remarkable capabilities in modeling complex distributions of natural images and videos. A pioneering work Stylelight \cite{stylelight} trains a dual-StyleGAN to simultaneously synthesize LDR and HDRI maps from the noisy latent, using GAN inversion to find the latent of the panorama corresponding to the input images during inference. Similarly, EverLight \cite{dastjerdi2023everlight} conditions a GAN on both input images and extracted parametric lighting to predict illumination for both indoor and outdoor scenes. More recently, DiffusionLight \cite{Phongthawee_2024_CVPR} has been proposed to inpaint a chrome ball as a light probe into the input image by fine-tuning a pre-trained 2D text-to-image diffusion model and controlling the exposure using the text condition. These multi-exposure balls are then warped to LDR panoramas and fused into an HDRI map. However, the visual quality of the result is affected by the ball-to-panorama warping, and the dynamic range is limited by the HDR fusion process. Another notable approach, LuxDiT \cite{liang2025luxdit}, fine-tunes a video diffusion transformer to directly predict scene lighting from an image or video to enhance visual quality. Although the above methods benefit from the strong priors of generative models, their training pipelines largely adhere to the conventional paradigm and lack a comprehensive spatial understanding. Since lighting estimation requires reasoning about the distribution of illumination and the scene layout, how to endow generative models with stronger awareness of both lighting and visual space remains underexplored.

\paragraph{Spatially-varying Lighting Estimation.} To extend beyond global illumination, some methods aim to recover the spatially-varying illumination. An intuitive approach is to estimate the illumination at the location of each pixel. These methods typically recover scene geometry from visual inputs and warp the scene into a partial panorama at a specific location, subsequently predicting a complete HDRI map from the warped version \cite{song2019neural,pointar,zhu2021spatially}.
Alternatively, other approaches directly regress the illumination at each pixel represented by spherical harmonics \cite{garon2019fast} or spherical Gaussians \cite{li2020inverse} using an end-to-end network. 
Moreover, instead of estimating 2D HDR lighting, some techniques aim to reconstruct spatially coherent 3D light fields using volumetric representations \cite{srinivasan2020lighthouse, Wang_2021_ICCV, li2023spatiotemporally} or voxel octrees \cite{Wang_2024_CVPR}. More recently, 4DLighting \cite{4dlighting} optimizes a 4D MLP light field using a 2D diffusion prior via score distillation sampling (SDS) to achieve spatiotemporally consistent lighting estimation. However, it is limited to monocular inputs and often produces over-smoothed results due to its heavy reliance on SDS. A concurrent work, LIMO \cite{Bolduc_2026_CVPR}, also enhances the diffusion model’s spatial awareness by providing richer geometric conditions, thereby enabling more accurate prediction of spatially varying illumination.

\begin{figure*}[ht]
    \centering
    \includegraphics[width=\linewidth]{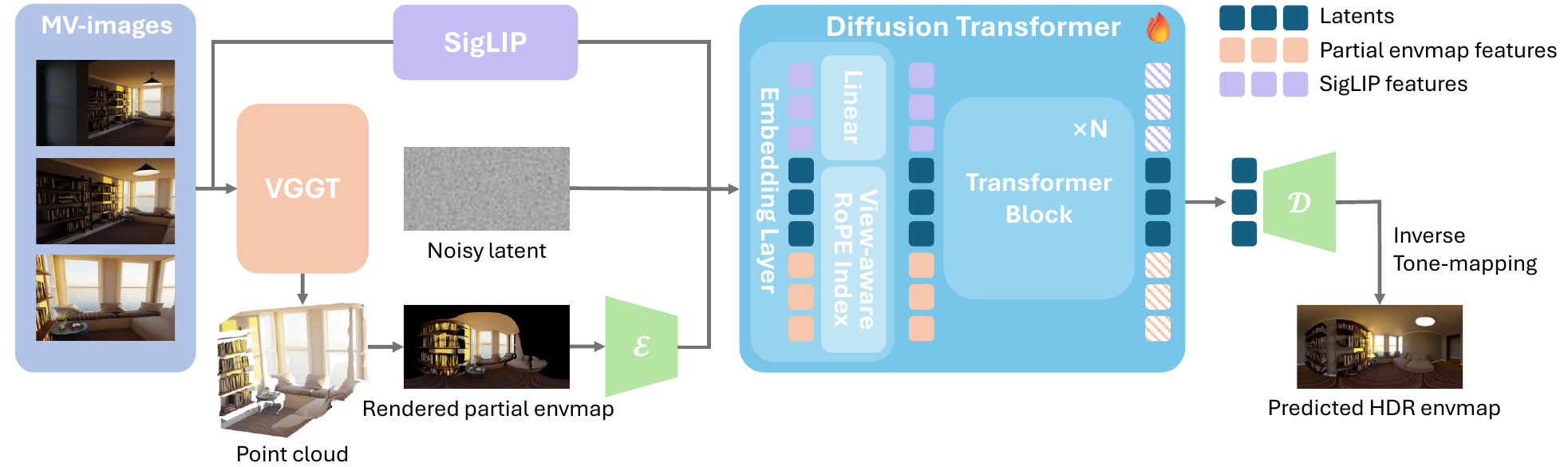} 
    \vspace{-17pt}
    \caption{Overview of LUMEN. Given a single-view image or multi-view inputs, we first estimate the camera parameters and reconstruct a coarse point cloud using an off-the-shelf feed-forward 3D reconstruction model. We then render partial environment maps at arbitrary locations in the scene based on the reconstructed geometry, and condition a 2D diffusion model on the input views with these partial environment maps to generate a complete HDRI map.}
    \vspace{-10pt}
    \label{fig:overview}
\end{figure*}

\paragraph{Inverse Rendering.} In contrast to methods that focus solely on estimating illumination from visual inputs, inverse rendering aims to jointly model and recover the scene geometry, materials, and illumination. Optimization-based inverse rendering methods \cite{srinivasan2021nerv, zhang2021nerfactor, GSIR, GIGS} utilize differentiable rendering to optimize the scene representation and illumination through photometric loss, supervised by dense multi-view captures. Nonetheless, these methods often simplify the optimization process by representing scene lighting as a global environment map, which lacks spatial awareness and fails to accurately characterize the complex light distribution in the scene. Some methods attempt to model the light fields that permeate through the space \cite{yao2022neilf, cache}, but the results are over-smoothed. On the other hand, learning-based methods \cite{li2020inverse, wang2021learning, zhu2022learning} use neural networks to directly predict scene attributes and illumination. 
However, these methods often rely on screen space information, lacking complete simulation of light-matter interactions in the scene, and thus struggle to accurately handle complex light transport and effects.

\section{Method}
In this section, we present our LUMEN framework for estimating both global and spatially-varying illumination, represented as HDRI maps, from single-view or multi-view image captures. Our key idea is to enhance the spatial awareness of the model by incorporating multi-view priors through both model design and dataset curation. This enables the model to predict the illumination of unseen regions in the scene. Specifically, we repurpose a pre-trained 2D diffusion transformer to generate HDRI maps conditioned on input image(s) and a partial LDR environment map rendered from a reconstructed point cloud (Sec~\ref{Lumen}). We further build a large-scale dataset that includes spatially-varying illumination from synthetic scenes and photo-realistic global illumination from real-world HDRI maps (Sec~\ref{data}). An overview of our framework is depicted in Fig.~\ref{fig:overview}.

\subsection{Lighting Prediction from Visual Captures}
\label{Lumen}
Our method predicts both global and spatially-varying lighting from visual inputs. To elucidate the overall framework, we first outline how spatially-varying lighting is inferred. Given a set of multi-view images $I\in \mathbb{R}^{3\times H\times W \times N}$, where $N$ is the number of images (with $N=1$ for a single-view image input), we aim to predict the HDRI map $H \in \mathbb{R}^{3\times H^{\prime}\times W^{\prime}}$ at a specific position $\boldsymbol{x}=(x,y,z)$. Since the scene geometry and camera parameters of the input views remain unknown, we use VGGT, an off-the-shelf feed-forward geometry reconstruction network \cite{VGGT}, to estimate the depth map $D$, camera intrinsics $K$, and camera extrinsics $[R|t]$ for each view. Subsequently, we back-project the depth points from each view to obtain the scene point cloud. Then, we render a partial LDR panorama $H_p \in \mathbb{R}^{3\times H^{\prime}\times W^{\prime}}$ from the point cloud by positioning an equirectangular camera at $\boldsymbol{x}$. The input multi-view images $I$ and this partial panorama $H_p$ are then sent to the 2D diffusion transformer as conditions for inferring the HDRI map.

\paragraph{Model Design} The Diffusion Transformer~\cite{dit}  has achieved significant success in image and video generation and perception, owing to its pure attention design that facilitates efficient information fusion across different inputs. In our framework, the DiT is conditioned on the input multi-view images, which provide visual and semantic annotations of the scene and the partial environment map that ensures the consistency between predicted results and the scene. However, directly concatenating condition tokens with latent tokens, as in previous work \cite{liang2025luxdit}, may lead to suboptimal performance, since perspective and panorama images perform different 2D spatial formulations. This difference makes it difficult for the pre-trained model to fuse multi-view information into latent tokens during the denoising process, as it lacks 3D perceptual capabilities. Therefore, we employ different encoders to extract features from different conditions, followed by an embedding layer that integrates these features into a unified token for processing by the transformer blocks. Moreover, since the value range of HDR images $H$ spans $[0,+\infty]$, tone-mapping is necessary to rescale the input to $[0,1]$. Previous methods have utilized both Reinhard~\cite{10.1145/566570.566575} and normalized log tone-mapping to preserve the high intensity in bright regions while maintaining details in dark areas. However, we find that only using Reinhard tone-mapping achieves comparable results, yielding the tone-mapped image $\boldsymbol{H}_L$ as follows:
\begin{equation}
   H_L = \frac{H}{1+H}.
\end{equation}

For input multi-view images, 
we use the SigLIP \cite{siglip} encoder to extract semantic features that guide the DiT to generate results with correct layout, color temperature, and style. The extracted features provide a rich and faithful representation of the scene, which we then project into VAE's latent space using a learnable linear layer, resulting in the multi-view feature $f_I=\text{Linear}\big[\text{SigLIP}(I)\big]\in \mathbb{R}^{N \times h \times w \times c}$. Similarly, the features of the partial environment map are extracted using a pre-trained VAE encoder, yielding $f_{H_p} = \text{VAE}(H_p) \in \mathbb{R}^{1 \times h^{\prime} \times w^{\prime} \times c}$ to accurately preserve visual information. Finally, the encoded multi-view images, the partial environment map, and the noisy latent $z_L$ are concatenated to form a unified input sequence of tokens. We also implement a view-aware RoPE \cite{su2024roformer} index on these unified tokens to help the model understand the spatial relationships between the noisy latent and the condition images. For each token at the 2D location $(i, j)$ in the $t$-th input (e.g., a multi-view image, the partial panorama, or the noisy latent), we first construct a RoPE index that places tokens from all inputs into a shared indexing space. Specifically, we define the RoPE index as follows:
\begin{equation}
\phi(i,j,t) = (i,j,0)+\sum_{k=1}^{t}(1,0,o_k),   
\end{equation}
where $o_k$ is a resolution-dependent offset associated with the $k$-th input. The accumulated offset separates different inputs along the indexing space, preventing index collisions and controlling inter-view separations. The unified sequence of tokens and the view-aware index are then sent to the DiT to produce the denoised latent:
\begin{equation}
\hat{z}_L = \text{DiT}\left(\left[f_I; f_{H_p};z \right] , \phi\right), 
\end{equation}
where $[;]$ denotes concatenation along the token dimension. Then, $\hat{z}_L$ is decoded using the VAE decoder to generate an LDR panorama, followed by inverse tone-mapping to recover the HDRI map. The model is trained using a standard flow-matching loss as follows:
\begin{equation}
\mathcal{L}=\mathbb{E}_{z_0, t, \epsilon \sim \mathcal{N}(0, I)}\left\|v_\theta\left(z_t ; t, f_I, f_{H_p}\right)-\left(\epsilon-z_0\right)\right\|_2^2,
\end{equation}
where $z_t$ is the latent representation at denoising step $t$ and $\epsilon-z_0$ represents the ground-truth velocity.

\subsection{Dataset Curation}
\label{data}
We construct our training data by combining a large-scale synthetic dataset and real-world HDRI maps from public data sources. The synthetic dataset contains multi-view RGB images and spatially-varying illumination of diverse indoor scenes, enabling the model to learn complex lighting distributions and enhance spatial awareness. In contrast, training on real-world data mitigates a decrease in the visual quality of generated images and improves robustness against in-the-wild scenarios. The specific data strategy is outlined below.

\paragraph{Synthetic Spatially-varying Lighting} Training our model requires a dataset with ground-truth spatially-varying lighting from various positions in the scene and the corresponding multi-view RGB images. Unfortunately, current public datasets with spatially-varying illumination cannot meet our requirements. For example, InterioNet \cite{li2018interiornet} provides only LDR panoramas, while OpenRooms \cite{Li_2021_CVPR} contains per-pixel spatially-varying illumination only on the surface of the scene. Among existing datasets, InteriorVerse \cite{zhu2022learning} is the only one that consists of true HDR spatially-varying illumination and multi-view RGB images for diverse indoor scenes, although the resolutions of its rendered panoramas are relatively low. To address these limitations, we leverage Infinigen Indoors \cite{Raistrick_2024_CVPR}, a rule-based framework, to generate 700 diverse indoor scenes with various layouts, material textures, and illumination. To enhance the diversity and controllability of the generation process, we incorporate 200 additional BSDF materials into the existing Infinigen Indoors material library and introduce extra rules to ensure that the generated room layouts align with predefined shapes, such as rectangles, convex hull, or L-shapes, facilitating subsequent camera position sampling. For each scene, we randomly generate a sky texture or sample an environment map from the selected 557 HDRI maps as the background in Blender, and adjust the strength to ensure reasonable brightness.

For each indoor scene, we place 12 cameras on a full orbit around the scene, each directed toward the center, while modulating their height and elevation using a sinusoid function. We also enforce a clear line of sight within a fixed forward-distance threshold to avoid occlusions and maximize scene coverage. Then, we randomly sample light probes in the scene and camera frustums. To be specific, we first extract the bounding convex hull of the scene and uniformly sample $[6,9]$ light probes, ensuring that no occluding objects are presented within proximity by performing ray intersection tests. In addition, we sample $[1,3]$ light probes in each camera's frustum, ensuring visibility from the camera's perspective without obstruction.

\paragraph{Real-world HDRI maps} Training our model using real-world HDR data enhances the generalizability and robustness of our method. We generate training pairs of 1,527 HDRI maps collected from four public data sources. Each training pair consists of an HDRI map and a limited field-of-view LDR cropped image obtained by projecting the panorama to a field-of-view image using a perspective camera. The perspective camera is generated with randomly sampled azimuth angles in $[0,360]$ degrees and elevation angles in $[-20,20]$ degrees, with a field of view that varies from 50 to 80 degrees. Then, we randomly select tone-mapping functions such as Gamma2.4, Flimic, and ACES to map the cropped images to the LDR format without exposure compensation.

\section{Experiments}
\subsection{Experiment Settings}
\paragraph{Implementation Details} We initialize our DiT from FLUX.1-dev \cite{blackforest2024flux} and fine-tune it using LoRA \cite{hu2022lora}. The resolutions of input images vary from $512\times 384$ to $768 \times768$, while the resolutions for point cloud rendering and output environment maps are set to $1024 \times 512$. We train our model for 20,000 steps using the AdamW optimizer \cite{kingma2014adam}, with a batch size of 8 and a warm-up phase of 2,000 steps on 8 NVIDIA H20 GPUs.

\paragraph{Datasets} We evaluate \name{} on five benchmark datasets encompassing diverse indoor/outdoor environments and featuring both global illumination and spatially-varying lighting. (1) Laval Indoor \cite{gardner2017learning}: following prior works \cite{Phongthawee_2024_CVPR, stylelight}, we use the same 289 HDRI maps from the official test split; (2) Laval Outdoor \cite{Hold-Geoffroy_2019_CVPR}: this dataset consists of 205 outdoor HDRI maps captured using a DSLR camera; (3) Laval Indoor Spatially Varying HDR Dataset \cite{garon2019fast}: this dataset consists of 20 indoor scenes with physical probes placed in them, captured using a DSLR camera; (4) Infinigen Indoors (single/multi-view): we construct a test set of 21 indoor scenes using Infinigen Indoor \cite{Raistrick_2024_CVPR}. For each scene, we: (i) sample one camera and place four light probes within its view frustum to render corresponding perspective images and HDRI maps for evaluating spatially-varying illumination from a single image; and (ii) additionally place $[6,9]$ cameras throughout the scene to evaluate performance with multi-view inputs; (5) In-the-wild: we collect 16 synthetic indoor scenes from the Internet and build the dataset using the same procedure as employed for Infinigen Indoors, but extend the amount of light probes to $[6,12]$.

\paragraph{Evaluation Protocol} We follow the three-sphere evaluation protocol adopted in prior works \cite{gardner2017learning, stylelight, Phongthawee_2024_CVPR}. Specifically, we use the predicted HDRI map (at a resolution of $128 \times 256$) to render three spheres with different materials (gray diffuse, silver matte, and silver mirror) and compare the resultant renderings with those generated using the ground-truth HDRI map. For global illumination estimation, we report scale-invariant RMSE (si-RMSE) \cite{grosse2009ground} and normalized RMSE (computed after mapping the 0.1st and 99.9th percentiles to 0 and 1) \cite{Phongthawee_2024_CVPR} to assess the predicted illumination distribution and structure, as well as angular error \cite{legendre2019deeplight} to evaluate the chromatic accuracy. For spatially-varying illumination, we replace normalized RMSE with RMSE to more accurately reflect intensity errors in the predicted illuminance. We also report the SSIM to evaluate structural fidelity. All metrics are averaged across all sample locations in all scenes.

\paragraph{Baselines} 
We compare our method against a comprehensive set of lighting estimation baselines to demonstrate its effectiveness. For global illumination estimation from a cropped image, we include StyleLight \cite{stylelight}, DiffusionLight \cite{Phongthawee_2024_CVPR}, and LuxDiT \cite{liang2025luxdit} as other methods do not release official inference code. Since StyleLight is trained on indoor data, we fine-tune it using the Laval Outdoor dataset to ensure a fair comparison. For spatially-varying illumination estimation from a single-view image, we compare against DiffusionLight \cite{Phongthawee_2024_CVPR} and 4DLighting \cite{4dlighting}. For spatially-varying illumination estimation from multi-view images, we adapt the inference pipeline of 4DLighting to accommodate multi-view inputs and compare our method with this extended baseline.

\begin{table*}[!t]
\centering
\large
\caption{Quantitative results on the Laval Indoor and Laval Outdoor datasets, underlined entries indicate the second-best results.}
\vspace{-9pt}
\label{tab:three_spheres}
\setlength{\tabcolsep}{6.8pt}
\renewcommand{\arraystretch}{0.8}
\begin{tabular}{
    l
    l
    c
    c
    c
    c
    c
    c
    c
    c
    c
}
\toprule
\multirow{2}{*}{\textbf{Dataset}} & \multirow{2}{*}{\textbf{Method}} & \multicolumn{3}{c}{\textbf{Scale-invariant RMSE} $\downarrow$} & \multicolumn{3}{c}{\textbf{Angular Error} $\downarrow$}  & \multicolumn{3}{c}{\textbf{Normalized RMSE} $\downarrow$}                                                            \\
 & & Diffuse           & Matte           & Mirror           & Diffuse           & Matte           & Mirror   & Diffuse           & Matte           & Mirror      \\ 
 [-2pt]
\midrule

\multirow{4}{*}{Laval Indoor}  
& StyleLight                       & 0.064 & 0.152 & \underline{0.265} & 3.874 & 4.606 & 10.974 & 0.232 & 0.411 & 0.520 \\
& DiffusionLight                   & 0.075 & 0.163 & 0.282 & 3.415 & 4.157 & 6.483 & 0.246 & 0.391 & \underline{0.457} \\
& LuxDiT                           & \underline{0.056} & \underline{0.140} & 0.274 & \underline{2.759} & \underline{3.875} & \textbf{6.099} & \underline{0.207} & \underline{0.388} & 0.546 \\
& Ours                             & \textbf{0.046} & \textbf{0.120} & \textbf{0.252} & \textbf{2.700} & \textbf{3.773} & \underline{6.257} & \textbf{0.180} & \textbf{0.319} & \textbf{0.411} \\
\hline
\\[-9pt]

\multirow{4}{*}{Laval Outdoor}
& StyleLight                       & 0.131 & 0.286 & 0.456 & 3.524 & 4.864 & 7.609 & 0.293 & 0.377 & 0.475 \\
& DiffusionLight                   & \underline{0.097} & \underline{0.252} & \underline{0.431} & \underline{2.838} & 4.237 & 6.835 & \underline{0.185} & \underline{0.320} & \underline{0.434} \\
& LuxDiT                           & 0.132 & 0.281 & 0.456 & \textbf{2.794} & \underline{4.149} & \underline{6.622} & 0.237 & 0.351 & 0.460 \\
& Ours                             & \textbf{0.087} & \textbf{0.221} & \textbf{0.403} & 2.940 & \textbf{3.898} & \textbf{6.309} & \textbf{0.162} & \textbf{0.297} & \textbf{0.423} \\
[-3pt]
\bottomrule

\end{tabular}
\vspace{-7pt}
\end{table*}

\begin{figure*}[h]
    \centering
    \includegraphics[width=1\linewidth]{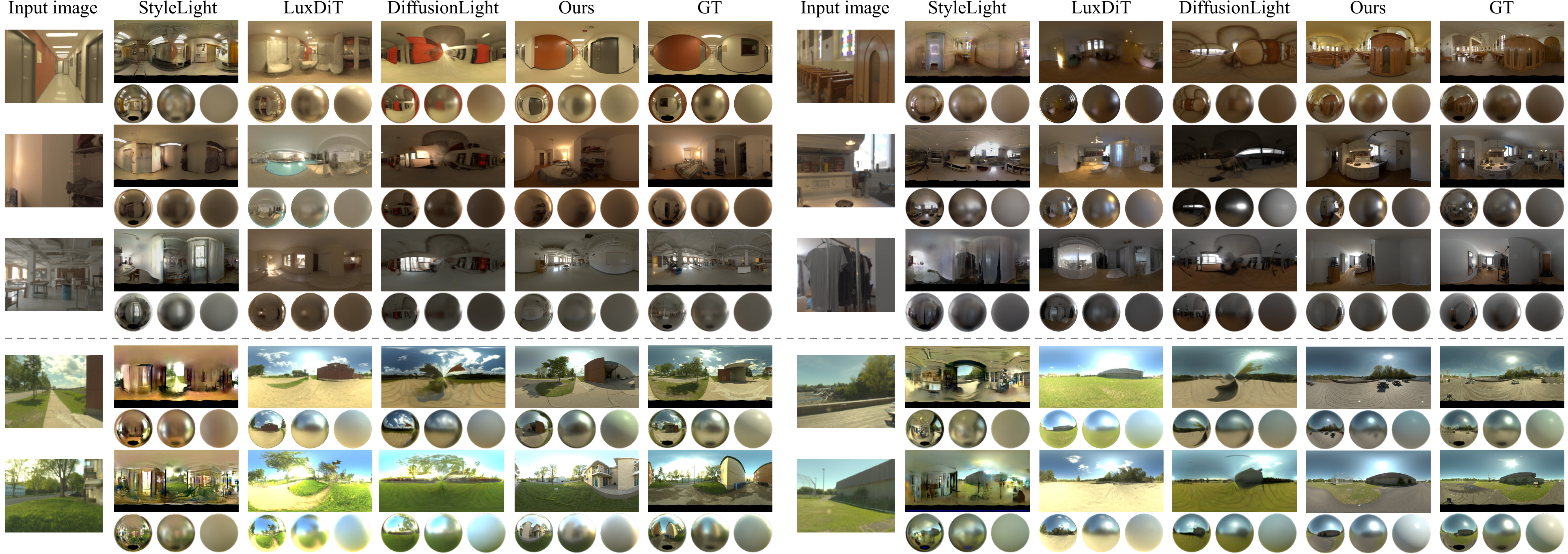} 
    \vspace{-15pt}
    \caption{Qualitative comparison on the Laval Indoor and Laval Outdoor datasets. (Please zoom in to see details)}
    \vspace{-12pt}
    \label{fig:GI}
\end{figure*}

\subsection{Global Illumination Estimation from a Single Image} We first evaluate our approach against prior generative methods for global illumination estimation from a cropped image. Table \ref{tab:three_spheres} reports quantitative results for both indoor and outdoor scenes. Our method outperforms previous approaches on the Laval Indoor and Laval Outdoor datasets across nearly all metrics. The only exception is the angular error for the rendered diffuse sphere. Nevertheless, our method still achieves higher accuracy in overall color reproduction, indicating that it effectively captures the scene’s illumination distribution with faithful chromaticity.

We further present qualitative comparisons in Fig. \ref{fig:GI}. Our method yields higher visual fidelity and more accurate localization of light sources in both indoor and outdoor environments. Both DiffusionLight and LuxDiT predict HDRI maps with plausible overall lighting distributions, but they often fail to capture fine-grained details. In addition, DiffusionLight exhibits noticeable distortions near the center of the HDRI map, likely stemming from its reliance on back-projection from a probe sphere. Moreover, StyleLight sometimes misplaces light sources in indoor scenes and fails to produce meaningful predictions for outdoor environments. The object insertion results in Fig. \ref{fig:object insertion} show that our method yields more accurate color, brightness, and shadowing than the baselines, suggesting enhanced estimation of illumination distribution, chromaticity, and intensity. In addition, we also provide image-conditioned HDRI maps synthesis results in the Appendix.
\begin{table*}[!t]
\centering
\small
\caption{Quantitative comparison for spatially-varying lighting estimation. The results are reported in terms of RMSE, si-RMSE, angular error, and SSIM.}
\vspace{-10pt}
\label{tab:sv}
\renewcommand{\arraystretch}{0.9} 
\resizebox{1\textwidth}{!}{%
\setlength{\tabcolsep}{1pt}
\setlength{\aboverulesep}{0pt}
\begin{tabular}{
    l@{\hspace{6pt}}
    l@{\hspace{4pt}}
    c@{\hspace{4pt}}c@{\hspace{4pt}}c@{\hspace{4pt}}
    c@{\hspace{4pt}}c@{\hspace{4pt}}c@{\hspace{4pt}}
    c@{\hspace{4pt}}c@{\hspace{4pt}}c@{\hspace{4pt}}
    c@{\hspace{4pt}}c@{\hspace{4pt}}c
}
\toprule
\multirow{2}{*}{\textbf{Scene}} & \multirow{2}{*}{\textbf{Method}} & \multicolumn{3}{c}{\textbf{RMSE} $\downarrow$} & \multicolumn{3}{c}{\textbf{si-RMSE} $\downarrow$} & \multicolumn{3}{c}{\textbf{Angular Error} $\downarrow$} & \multicolumn{3}{c}{\textbf{SSIM} $\uparrow$} \\
\cmidrule(lr){3-5} \cmidrule(lr){6-8} \cmidrule(lr){9-11} \cmidrule(lr){12-14}
 & & Diffuse & Matte & Mirror & Diffuse & Matte & Mirror & Diffuse & Matte & Mirror & Diffuse & Matte & Mirror \\ 
 [-2pt]
\midrule

\multirow{3}{*}{Infinigen (single)}  

& DiffusionLight (CVPR'24)          & 0.157 & 0.228 & 0.460 & 0.066 & 0.180 & 0.428 & 2.935 & 4.053 & 5.394 & 0.947 & 0.679 & 0.471 \\
& 4DLighting (SIGGRAPH'25)               & \underline{0.164} & \underline{0.207} & \underline{0.430} & \underline{0.043} & \underline{0.134} & \underline{0.381} & \textbf{1.457} & \textbf{2.418} & \textbf{3.595} & \underline{0.972} & \underline{0.754} & \underline{0.533} \\
& Ours                             & \textbf{0.110} & \textbf{0.149} & \textbf{0.369} & \textbf{0.042} & \textbf{0.108} & \textbf{0.329} & \underline{2.762} & \underline{3.382} & \underline{4.594} & \textbf{0.973} & \textbf{0.832} & \textbf{0.679} \\
\hline
\\[-7pt]

\multirow{3}{*}{In-the-wild (single)}
& DiffusionLight (CVPR'24)          & 0.172 & 0.236 & 0.351 & 0.081 & 0.193 & 0.310 & 1.592 & 2.681 & 3.856 & 0.923 & 0.654 & 0.447 \\
& 4DLighting (SIGGRAPH'25)               & \textbf{0.141} & \underline{0.190} & \underline{0.289} & \underline{0.048} & \underline{0.132} & \underline{0.238} & \textbf{1.186} & \textbf{1.904} & \textbf{2.865} & \underline{0.969} & \underline{0.755} & \underline{0.519} \\
& Ours                             & \underline{0.153} & \textbf{0.176} & \textbf{0.276} & \textbf{0.047} & \textbf{0.114} & \textbf{0.225} & \underline{1.345} & \underline{2.379} & \underline{3.563} & \textbf{0.964} & \textbf{0.797} & \textbf{0.616} \\

\hline
\\[-7pt]

\multirow{3}{*}{Laval Indoor SV (single)}
& DiffusionLight (CVPR'24)          & 0.262 & 0.289 & 0.365 & \underline{0.072} & \underline{0.169} & 0.253 & \underline{2.882} & \underline{3.604} & \underline{4.316} & \underline{0.931} & \underline{0.728} & 0.572 \\
& 4DLighting (SIGGRAPH'25)               & \textbf{0.169} & \underline{0.247} & \underline{0.322} & 0.085 & 0.180 & \underline{0.251} & 3.099 & 3.753 & 4.459 & 0.915 & 0.698 & \underline{0.577} \\
& Ours                             & \underline{0.183} & \textbf{0.216} & \textbf{0.298} & \textbf{0.069} & \textbf{0.154} & \textbf{0.248} & \textbf{2.572} & \textbf{3.493} & \textbf{4.167} & \textbf{0.939} & \textbf{0.769} & \textbf{0.599} \\

\hline
\\[-7pt]

\multirow{2}{*}{Infinigen (multi-view)}
& 4DLighting (SIGGRAPH'25)               & 0.213 & 0.231 & 0.443 & 0.037 & 0.124 & 0.375 & \textbf{1.596} & \textbf{2.397} & 3.568 & 0.978 & 0.762 & 0.511 \\
& Ours                             & \textbf{0.080} & \textbf{0.095} & \textbf{0.253} & \textbf{0.024} & \textbf{0.058} & \textbf{0.210} & 1.925 & 2.416 & \textbf{3.340} & \textbf{0.988} & \textbf{0.942} & \textbf{0.849} \\

\hline
\\[-7pt]

\multirow{2}{*}{In-the-wild (multi-view)}
& 4DLighting (SIGGRAPH'25)               & 0.182 & 0.204 & 0.291 & 0.038 & 0.111 & 0.210 & \textbf{1.168} & \textbf{1.784} & \textbf{2.728} & 0.972 & 0.773 & 0.505 \\
& Ours                             & \textbf{0.104} & \textbf{0.118} & \textbf{0.172} & \textbf{0.027} & \textbf{0.065} & \textbf{0.143} & 1.522 & 1.898 & 3.003 & \textbf{0.984} & \textbf{0.907} & \textbf{0.788} \\
[-1pt]
\bottomrule

\end{tabular}}
\vspace{-10pt}
\end{table*}

\begin{figure*}[h]
    \centering
    \includegraphics[width=0.99\linewidth]{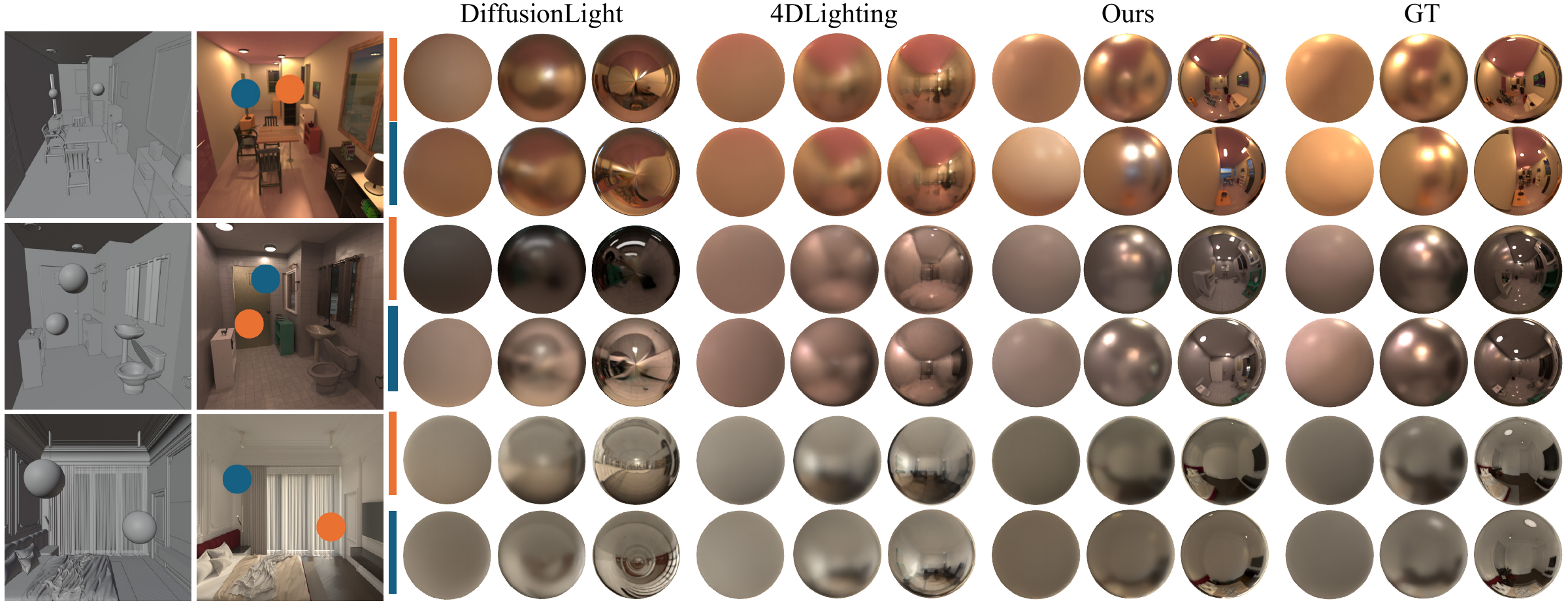} 
    \vspace{-10pt}
    \caption{Qualitative comparison of spatially-varying lighting estimation from a single-view image.}
    \vspace{-10pt}
    \label{fig:sv1}
\end{figure*}

\subsection{Spatially-varying Lighting Estimation} 
\begin{figure*}[h]
    \centering
    \includegraphics[width=0.99\linewidth]{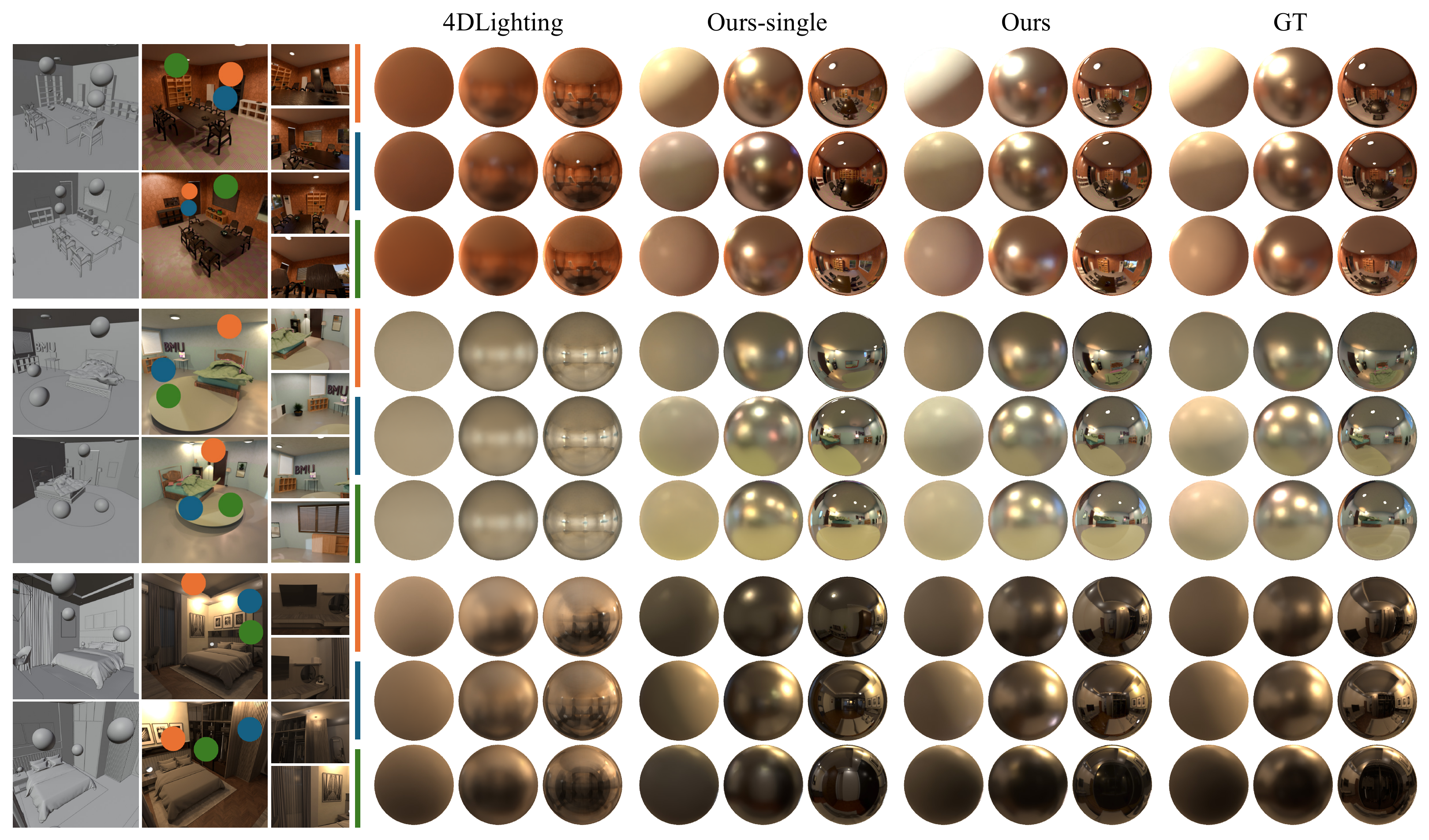} 
    \vspace{-10pt}
    \caption{Qualitative comparison of spatially-varying lighting estimation from multi-view images.}
    \vspace{-5pt}
    \label{fig:sv2}
\end{figure*}

To evaluate spatially-varying lighting estimation, we compare our method with representative baselines on the Laval Indoor Spatially Varying (SV), Infinigen Indoor, and In-the-wild datasets. For Infinigen Indoor and In-the-wild datasets, we report results under both single-view and multi-view inputs to assess how different methods benefit from additional visual observations. Table \ref{tab:sv} summarizes the quantitative comparisons, while Figs. \ref{fig:sv1} and \ref{fig:sv2} provide qualitative results. Our method outperforms the baselines in terms of RMSE, si-RMSE, and SSIM on two synthetic datasets, achieving more accurate estimation of illumination intensity and spatial distribution, and is only surpassed by 4DLighting in angular error. We note that 4DLighting is trained exclusively on synthetic scenes generated by Infinigen Indoor and rendered in Blender, and thus owns a particularly strong dataset-specific prior of tone and chromaticity. Furthermore, our method consistently outperforms all baselines on every metric over the real-world Laval Indoor SV dataset, highlighting its robustness and generalizability.

\begin{figure*}[h]
    \centering
    \includegraphics[width=0.99\linewidth]{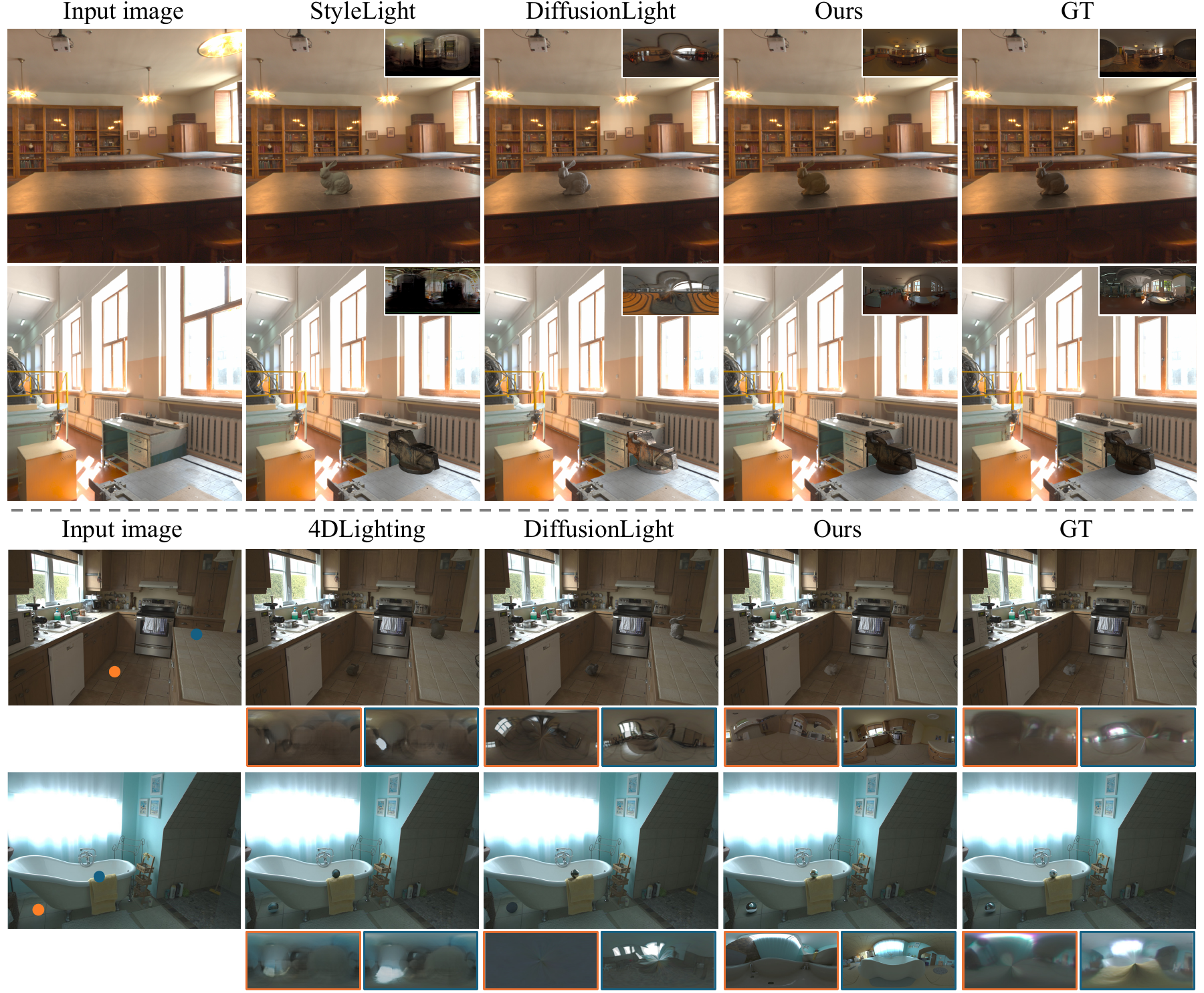} 
    \vspace{-10pt}
    \caption{Qualitative comparison of object insertion on the Laval Indoor dataset (top) and the Laval Indoor Spatially Varying HDR dataset (below). (Please zoom in to see details)}
    \vspace{-5pt}
    \label{fig:object insertion}
\end{figure*}

For spatially-varying lighting estimation from a single-view image, the results in Fig. \ref{fig:sv1} show that our method accurately localizes light sources, predicts their intensities, and reconstructs spatially coherent illumination across various probe locations. In contrast, DiffusionLight, which is primarily trained for global illumination estimation, yields inconsistent predictions and mislocalized light sources in the spatially-varying context. While 4DLighting produces spatially coherent results, its predictions show marginal variation as the probe location changes, and it may inaccurately place light sources. Moreover, the over-smoothed nature of its MLP-based light-field representation often leads to a lack of concentrated high-intensity light sources. The object insertion results in Fig. \ref{fig:object insertion} further shows that our method faithfully recovers real-world spatially-varying illumination, whereas the baseline methods fail to do so.

\begin{figure*}[h]
    \centering
    \includegraphics[width=0.99\linewidth]{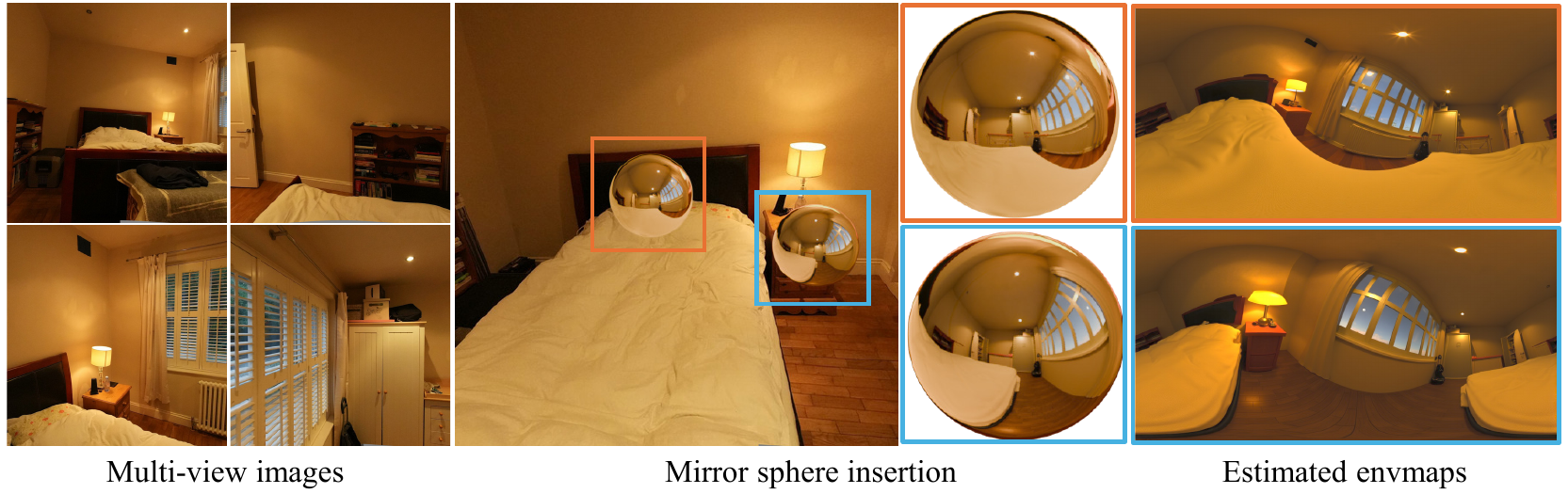} 
    \vspace{-10pt}
    \caption{Visualizations of object insertion and the estimated HDRI maps from multi-view input images.}
    \vspace{-5pt}
    \label{fig:real}
\end{figure*}

With multi-view inputs, Table \ref{tab:sv} and Fig. \ref{fig:sv2} show that our method consistently benefits from the additional information, yielding more accurate light source localization and improved chromaticity compared to the single-view setting. In contrast, 4DLighting exhibits limited gains, with the results in Fig. \ref{fig:sv2} revealing artifacts and periodic patterns. This suggests that it does not reliably maintain consistent predictions across viewpoints as the camera pose changes. We also provide more real-world results in Fig. \ref{fig:real} and the Appendix.

\begin{table*}[h]
\centering
\footnotesize
\caption{Ablation results on Laval indoor and in-the-wild scenes. Results are reported in terms of RMSE, si-RMSE, angular error, and normalized RMSE.}
\vspace{-10pt}
\label{tab:ablation}
\renewcommand{\arraystretch}{0.8}
\resizebox{1\textwidth}{!}{%
\setlength{\tabcolsep}{1.5pt}
\begin{tabular}{
    l@{\hspace{6pt}}
    l@{\hspace{4pt}}
    c@{\hspace{4pt}}c@{\hspace{4pt}}c@{\hspace{4pt}}
    c@{\hspace{4pt}}c@{\hspace{4pt}}c@{\hspace{4pt}}
    c@{\hspace{4pt}}c@{\hspace{4pt}}c@{\hspace{4pt}}
    c@{\hspace{4pt}}c@{\hspace{4pt}}c
}
\toprule
\multirow{2}{*}{\textbf{Scene}} & \multirow{2}{*}{\textbf{Method}} & \multicolumn{3}{c}{\textbf{RMSE} $\downarrow$} & \multicolumn{3}{c}{\textbf{si-RMSE} $\downarrow$} & \multicolumn{3}{c}{\textbf{Angular Error} $\downarrow$} & \multicolumn{3}{c}{\textbf{N-RMSE} $\downarrow$} \\
\cmidrule(lr){3-5} \cmidrule(lr){6-8} \cmidrule(lr){9-11} \cmidrule(lr){12-14}
 & & Diffuse & Matte & Mirror & Diffuse & Matte & Mirror & Diffuse & Matte & Mirror & Diffuse & Matte & Mirror \\
[-2pt]
\midrule

\multirow{4}{*}{Laval Indoor}
& Single-only     & 0.268 & 0.287 & 0.409 & 0.053 & 0.130 & 0.259 & \underline{3.654} & \underline{4.359} & \underline{6.436} & 0.231 & 0.396 & 0.508 \\
& Infinigen-only  & 0.304 & 0.318 & 0.458 & \underline{0.052} & \underline{0.128} & 0.264 & 3.662 & 4.545 & 6.826 & 0.197 & 0.401 & 0.557 \\
& w/o SigLIP  & \underline{0.202} & \underline{0.242} & \underline{0.362} & 0.058 & 0.129 & \textbf{0.249} & 4.776 & 5.790 & 8.743 & \underline{0.195} & \underline{0.347} & \underline{0.450} \\
& Ours       & \textbf{0.184} & \textbf{0.218} & \textbf{0.351} & \textbf{0.046} & \textbf{0.120} & \underline{0.252} & \textbf{2.700} & \textbf{3.773} & \textbf{6.257} & \textbf{0.180} & \textbf{0.319} & \textbf{0.411} \\
[-1pt]
\midrule

\multirow{4}{*}{In-the-wild}

& Single-only     & 0.502 & 0.457 & 0.503 & 0.056 & 0.123 & 0.200 & 2.525 & 2.975 & 3.619 & 0.228 & 0.543 & 0.647 \\
& Infinigen-only  & 0.130 & 0.138 & 0.227 & \underline{0.031} & 0.079 & 0.192 & \underline{1.906} & \underline{2.071} & \textbf{2.854} & 0.114 & 0.225 & 0.296 \\
& w/o SigLIP  & \textbf{0.103} & \underline{0.118} & \underline{0.204} & 0.034 & \underline{0.077} & \underline{0.176} & 2.330 & 2.703 & 4.218 & \underline{0.111} & \underline{0.205} & \underline{0.269} \\
& Ours       & \underline{0.104} & \textbf{0.118} & \textbf{0.172} & \textbf{0.027} & \textbf{0.065} & \textbf{0.143} & \textbf{1.522} & \textbf{1.898} & \underline{3.003} & \textbf{0.092} & \textbf{0.172} & \textbf{0.225} \\
[-1pt]
\bottomrule
\end{tabular}}
\vspace{-10pt}

\end{table*}

\begin{figure*}[h]
    \centering
    \includegraphics[width=0.99\linewidth]{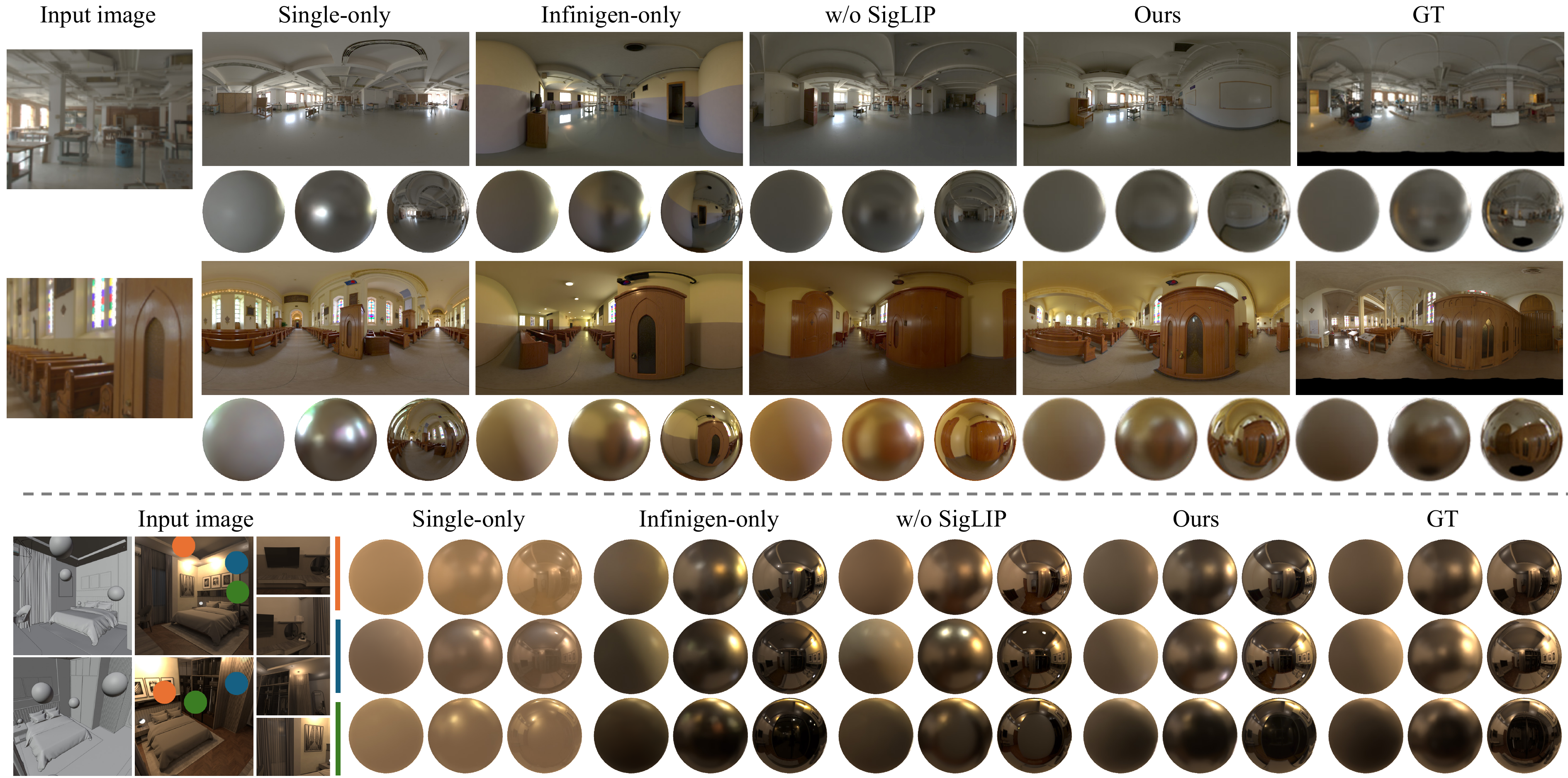} 
    \vspace{-10pt}
    \caption{Qualitative comparison of ablations.}
    \vspace{-5pt}
    \label{fig:ablation}
\end{figure*}

\subsection{Ablation Study} To investigate the impact of our architecture design and data strategy, we conduct an ablation study by training three variants of our model: (1) \emph{Single-only}, which is trained using paired cropped images and HDRI maps only from real-world datasets; (2) \emph{Infinigen-only}, which is trained solely on our synthetic dataset; and (3) \emph{w/o SigLIP}, in which we remove the SigLIP conditioning and the corresponding embedding layer. Table \ref{tab:ablation} reports quantitative results on Laval Indoor and in-the-wild spatially-varying datasets. Our model consistently outperforms all variants, validating the effectiveness of our design choices.

We further provide qualitative comparisons in Fig. \ref{fig:ablation}. The full model produces results that more closely align with the ground truth in terms of overall color accuracy, intensity, and illumination distribution. The \emph{Single-only} variant struggles to accurately localize and model light-source distributions for global illumination estimation and largely fails in spatially varying lighting scenarios due to the lack of spatial supervision in its training data. The \emph{Infinigen-only} variant exhibits reduced visual fidelity, reflecting the domain gap between synthetic renders and real captures. Lastly, the \emph{w/o SigLIP} variant cannot faithfully recover scene layouts and chromaticity, indicating that SigLIP features provide useful semantic cues for conditioning the model on the input image.

\section{Conclusion}
In this paper, we introduce LUMEN, a generative approach for estimating both global illumination and spatially-varying lighting from either a single image or multiple views. Our method repurposes a pre-trained diffusion model to infer HDRI maps from visual observations, and improves its spatial awareness and understanding of lighting by incorporating multi-view priors through (i) an embedding layer that fuses multi-view features into unified tokens, and (ii) a large-scale synthetic dataset containing diverse spatially-varying scenes and illumination conditions. Extensive experiments show that LUMEN consistently outperforms prior methods in both global environment-map estimation and spatially-varying lighting estimation, while producing photorealistic results.

\paragraph{Limitations and Discussion} Our approach relies on the learned prior and spatial reasoning capabilities of a 2D diffusion model to enforce consistency in spatially-varying illumination. In an ideal setting, given a large-scale, high-quality dataset of spatially varying illumination, our model can learn the complex spatial distribution of illumination. With sufficiently informative inputs, such as multi-view images, the model can faithfully recover the spatially varying illumination in a scene. However, when the input views cover only a limited portion of the scene, the model samples from the learned distribution for unobserved regions and produces plausible results. In this case, it cannot guarantee that the predicted illumination is consistent with the ground truth. In this sense, the task is closer to geometry-informed generation.

Besides, our method uses only Reinhard tone mapping to rescale HDR images. Although our experimental results (see in Appendix) show that this is sufficient in most cases, Reinhard tone mapping may fail to preserve the dynamic range of the input under extreme illumination conditions, such as direct sunlight in outdoor environments or sharp indoor spotlights. This can lead to numerical instability in the predicted light-source intensities, as well as chromatic deviations. For such scenes, incorporating normalized log tone mapping in addition to Reinhard tone mapping can more effectively recover light-source information. However, predicting results under both tone-mapping operators introduces additional computational overhead and requires extra mechanisms to ensure consistency between the two outputs, potentially calling for further model-design considerations. Therefore, we believe that the appropriate design choice should depend on the intended application. Since our work focuses on general scenes, we use only Reinhard tone mapping.

Moreover, our model does not explicitly address spatiotemporal illumination changes. We expect that extending our framework to video diffusion models could alleviate these limitations by leveraging their inherent spatiotemporal coherence.

\clearpage
\bibliographystyle{ACM-Reference-Format}
\bibliography{sample-base}

@String{Computing = "Computing" }

@String{Computer = "{IEEE} Computer" }

@String{Springer = "Springer-Verlag" }

@article{gardner2017learning,
  title={Learning to predict indoor illumination from a single image},
  author={Gardner, Marc-Andr{\'e} and Sunkavalli, Kalyan and Yumer, Ersin and Shen, Xiaohui and Gambaretto, Emiliano and Gagn{\'e}, Christian and Lalonde, Jean-Fran{\c{c}}ois},
  journal={ACM Transactions on Graphics (TOG)},
  volume={36},
  number={6},
  pages={1--14},
  year={2017},
  publisher={ACM New York, NY, USA}
}

@inproceedings{legendre2019deeplight,
  title={Deeplight: Learning illumination for unconstrained mobile mixed reality},
  author={LeGendre, Chloe and Ma, Wan-Chun and Fyffe, Graham and Flynn, John and Charbonnel, Laurent and Busch, Jay and Debevec, Paul},
  booktitle={Proceedings of the IEEE/CVF conference on computer vision and pattern recognition},
  pages={5918--5928},
  year={2019}
}

@inproceedings{zhan2021emlight,
  title={Emlight: Lighting estimation via spherical distribution approximation},
  author={Zhan, Fangneng and Zhang, Changgong and Yu, Yingchen and Chang, Yuan and Lu, Shijian and Ma, Feiying and Xie, Xuansong},
  booktitle={Proceedings of the AAAI Conference on Artificial Intelligence},
  volume={35},
  number={4},
  pages={3287--3295},
  year={2021}
}

@inproceedings{dastjerdi2023everlight,
  title={Everlight: Indoor-outdoor editable hdr lighting estimation},
  author={Dastjerdi, Mohammad Reza Karimi and Eisenmann, Jonathan and Hold-Geoffroy, Yannick and Lalonde, Jean-Fran{\c{c}}ois},
  booktitle={Proceedings of the IEEE/CVF international conference on computer vision},
  pages={7420--7429},
  year={2023}
}

@inproceedings{srinivasan2020lighthouse,
  title={Lighthouse: Predicting lighting volumes for spatially-coherent illumination},
  author={Srinivasan, Pratul P and Mildenhall, Ben and Tancik, Matthew and Barron, Jonathan T and Tucker, Richard and Snavely, Noah},
  booktitle={Proceedings of the IEEE/CVF Conference on Computer Vision and Pattern Recognition},
  pages={8080--8089},
  year={2020}
}

@article{li2023spatiotemporally,
  title={Spatiotemporally consistent hdr indoor lighting estimation},
  author={Li, Zhengqin and Yu, Li and Okunev, Mikhail and Chandraker, Manmohan and Dong, Zhao},
  journal={ACM Transactions on Graphics},
  volume={42},
  number={3},
  pages={1--15},
  year={2023},
  publisher={ACM New York, NY, USA}
}

@inproceedings{garon2019fast,
  title={Fast spatially-varying indoor lighting estimation},
  author={Garon, Mathieu and Sunkavalli, Kalyan and Hadap, Sunil and Carr, Nathan and Lalonde, Jean-Fran{\c{c}}ois},
  booktitle={Proceedings of the IEEE/CVF Conference on Computer Vision and Pattern Recognition},
  pages={6908--6917},
  year={2019}
}

@inproceedings{song2019neural,
  title={Neural illumination: Lighting prediction for indoor environments},
  author={Song, Shuran and Funkhouser, Thomas},
  booktitle={Proceedings of the IEEE/CVF Conference on Computer Vision and Pattern Recognition},
  pages={6918--6926},
  year={2019}
}

@inproceedings{li2020inverse,
  title={Inverse rendering for complex indoor scenes: Shape, spatially-varying lighting and svbrdf from a single image},
  author={Li, Zhengqin and Shafiei, Mohammad and Ramamoorthi, Ravi and Sunkavalli, Kalyan and Chandraker, Manmohan},
  booktitle={Proceedings of the IEEE/CVF conference on computer vision and pattern recognition},
  pages={2475--2484},
  year={2020}
}

@inproceedings{zhu2021spatially,
  title={Spatially-varying outdoor lighting estimation from intrinsics},
  author={Zhu, Yongjie and Zhang, Yinda and Li, Si and Shi, Boxin},
  booktitle={Proceedings of the IEEE/CVF conference on computer vision and pattern recognition},
  pages={12834--12842},
  year={2021}
}

@InProceedings{Phongthawee_2024_CVPR,
    author    = {Phongthawee, Pakkapon and Chinchuthakun, Worameth and Sinsunthithet, Nontaphat and Jampani, Varun and Raj, Amit and Khungurn, Pramook and Suwajanakorn, Supasorn},
    title     = {DiffusionLight: Light Probes for Free by Painting a Chrome Ball},
    booktitle = {Proceedings of the IEEE/CVF Conference on Computer Vision and Pattern Recognition (CVPR)},
    month     = {June},
    year      = {2024},
    pages     = {98-108}
}

@misc{liang2025luxdit,
    title={LuxDiT: Lighting Estimation with Video Diffusion Transformer},
    author={Ruofan Liang and Kai He and Zan Gojcic and Igor Gilitschenski and Sanja Fidler 
        and Nandita Vijaykumar and Zian Wang},
    year={2025},
    eprint={2509.03680},
    archivePrefix={arXiv},
    primaryClass={cs.GR}
}

@inproceedings{4dlighting,
author = {Tong, Mutian and Wu, Rundi and Zheng, Changxi},
title = {Spatiotemporally Consistent Indoor Lighting Estimation with Diffusion Priors},
year = {2025},
isbn = {9798400715402},
publisher = {Association for Computing Machinery},
address = {New York, NY, USA},
url = {https://doi.org/10.1145/3721238.3730749},
doi = {10.1145/3721238.3730749},
booktitle = {Proceedings of the Special Interest Group on Computer Graphics and Interactive Techniques Conference Conference Papers},
articleno = {107},
numpages = {11},
location = {
},
series = {SIGGRAPH Conference Papers '25}
}

@InProceedings{Raistrick_2024_CVPR,
    author    = {Raistrick, Alexander and Mei, Lingjie and Kayan, Karhan and Yan, David and Zuo, Yiming and Han, Beining and Wen, Hongyu and Parakh, Meenal and Alexandropoulos, Stamatis and Lipson, Lahav and Ma, Zeyu and Deng, Jia},
    title     = {Infinigen Indoors: Photorealistic Indoor Scenes using Procedural Generation},
    booktitle = {Proceedings of the IEEE/CVF Conference on Computer Vision and Pattern Recognition (CVPR)},
    month     = {June},
    year      = {2024},
    pages     = {21783-21794}
}

@InProceedings{pointar,
author="Zhao, Yiqin
and Guo, Tian",
editor="Vedaldi, Andrea
and Bischof, Horst
and Brox, Thomas
and Frahm, Jan-Michael",
title="PointAR: Efficient Lighting Estimation for Mobile Augmented Reality",
booktitle="Computer Vision -- ECCV 2020",
year="2020",
publisher="Springer International Publishing",
address="Cham",
pages="678--693",
isbn="978-3-030-58592-1"
}

@InProceedings{Wang_2021_ICCV,
    author    = {Wang, Zian and Philion, Jonah and Fidler, Sanja and Kautz, Jan},
    title     = {Learning Indoor Inverse Rendering With 3D Spatially-Varying Lighting},
    booktitle = {Proceedings of the IEEE/CVF International Conference on Computer Vision (ICCV)},
    month     = {October},
    year      = {2021},
    pages     = {12538-12547}
}

@InProceedings{Wang_2024_CVPR,
    author    = {Wang, Xuecan and Xiao, Shibang and Liang, Xiaohui},
    title     = {LightOctree: Lightweight 3D Spatially-Coherent Indoor Lighting Estimation},
    booktitle = {Proceedings of the IEEE/CVF Conference on Computer Vision and Pattern Recognition (CVPR)},
    month     = {June},
    year      = {2024},
    pages     = {4536-4545}
}

@inproceedings{probe1,
author = {Debevec, Paul},
title = {Rendering synthetic objects into real scenes: bridging traditional and image-based graphics with global illumination and high dynamic range photography},
year = {1998},
isbn = {0897919998},
publisher = {Association for Computing Machinery},
address = {New York, NY, USA},
url = {https://doi.org/10.1145/280814.280864},
doi = {10.1145/280814.280864},
booktitle = {Proceedings of the 25th Annual Conference on Computer Graphics and Interactive Techniques},
pages = {189–198},
numpages = {10},
series = {SIGGRAPH '98}
}

@article{eyes,
author = {Nishino, Ko and Nayar, Shree K.},
title = {Eyes for relighting},
year = {2004},
issue_date = {August 2004},
publisher = {Association for Computing Machinery},
address = {New York, NY, USA},
volume = {23},
number = {3},
issn = {0730-0301},
url = {https://doi.org/10.1145/1015706.1015783},
doi = {10.1145/1015706.1015783},
journal = {ACM Trans. Graph.},
month = aug,
pages = {704–711},
numpages = {8}
}

@InProceedings{Yi_2018_ECCV,
author = {Yi, Renjiao and Zhu, Chenyang and Tan, Ping and Lin, Stephen},
title = {Faces as Lighting Probes via Unsupervised Deep Highlight Extraction},
booktitle = {Proceedings of the European Conference on Computer Vision (ECCV)},
month = {September},
year = {2018}
}

@InProceedings{chip,
author = {Park, Jeong Joon and Holynski, Aleksander and Seitz, Steven M.},
title = {Seeing the World in a Bag of Chips},
booktitle = {Proceedings of the IEEE/CVF Conference on Computer Vision and Pattern Recognition (CVPR)},
month = {June},
year = {2020}
}

@InProceedings{Hold-Geoffroy_2019_CVPR,
author = {Hold-Geoffroy, Yannick and Athawale, Akshaya and Lalonde, Jean-Francois},
title = {Deep Sky Modeling for Single Image Outdoor Lighting Estimation},
booktitle = {Proceedings of the IEEE/CVF Conference on Computer Vision and Pattern Recognition (CVPR)},
month = {June},
year = {2019}
}

@InProceedings{Gardner_2019_ICCV,
author = {Gardner, Marc-Andre and Hold-Geoffroy, Yannick and Sunkavalli, Kalyan and Gagne, Christian and Lalonde, Jean-Francois},
title = {Deep Parametric Indoor Lighting Estimation},
booktitle = {Proceedings of the IEEE/CVF International Conference on Computer Vision (ICCV)},
month = {October},
year = {2019}
}

@InProceedings{Zhang_2019_CVPR,
author = {Zhang, Jinsong and Sunkavalli, Kalyan and Hold-Geoffroy, Yannick and Hadap, Sunil and Eisenman, Jonathan and Lalonde, Jean-Francois},
title = {All-Weather Deep Outdoor Lighting Estimation},
booktitle = {Proceedings of the IEEE/CVF Conference on Computer Vision and Pattern Recognition (CVPR)},
month = {June},
year = {2019}
}

@inproceedings{LeGendre_SA_20,
author = {LeGendre, Chloe and Ma, Wan-Chun and Pandey, Rohit and Fanello, Sean and Rhemann, Christoph and Dourgarian, Jason and Busch, Jay and Debevec, Paul},
title = {Learning Illumination from Diverse Portraits},
year = {2020},
isbn = {9781450380805},
publisher = {Association for Computing Machinery},
address = {New York, NY, USA},
url = {https://doi.org/10.1145/3410700.3425432},
doi = {10.1145/3410700.3425432},
booktitle = {SIGGRAPH Asia 2020 Technical Communications},
articleno = {7},
numpages = {4},
location = {Virtual Event, Republic of Korea},
series = {SA '20}
}

@InProceedings{stylelight,
author="Wang, Guangcong
and Yang, Yinuo
and Loy, Chen Change
and Liu, Ziwei",
editor="Avidan, Shai
and Brostow, Gabriel
and Ciss{\'e}, Moustapha
and Farinella, Giovanni Maria
and Hassner, Tal",
title="StyleLight: HDR Panorama Generation for Lighting Estimation and Editing",
booktitle="Computer Vision -- ECCV 2022",
year="2022",
publisher="Springer Nature Switzerland",
address="Cham",
pages="477--492",
isbn="978-3-031-19784-0"
}

@InProceedings{VGGT,
    author    = {Wang, Jianyuan and Chen, Minghao and Karaev, Nikita and Vedaldi, Andrea and Rupprecht, Christian and Novotny, David},
    title     = {VGGT: Visual Geometry Grounded Transformer},
    booktitle = {Proceedings of the IEEE/CVF Conference on Computer Vision and Pattern Recognition (CVPR)},
    month     = {June},
    year      = {2025},
    pages     = {5294-5306}
}

@article{li2018interiornet,
  title={Interiornet: Mega-scale multi-sensor photo-realistic indoor scenes dataset},
  author={Li, Wenbin and Saeedi, Sajad and McCormac, John and Clark, Ronald and Tzoumanikas, Dimos and Ye, Qing and Huang, Yuzhong and Tang, Rui and Leutenegger, Stefan},
  journal={arXiv preprint arXiv:1809.00716},
  year={2018}
}

@InProceedings{Li_2021_CVPR,
    author    = {Li, Zhengqin and Yu, Ting-Wei and Sang, Shen and Wang, Sarah and Song, Meng and Liu, Yuhan and Yeh, Yu-Ying and Zhu, Rui and Gundavarapu, Nitesh and Shi, Jia and Bi, Sai and Yu, Hong-Xing and Xu, Zexiang and Sunkavalli, Kalyan and Hasan, Milos and Ramamoorthi, Ravi and Chandraker, Manmohan},
    title     = {OpenRooms: An Open Framework for Photorealistic Indoor Scene Datasets},
    booktitle = {Proceedings of the IEEE/CVF Conference on Computer Vision and Pattern Recognition (CVPR)},
    month     = {June},
    year      = {2021},
    pages     = {7190-7199}
}

@inproceedings{siglip,
  title={Sigmoid loss for language image pre-training},
  author={Zhai, Xiaohua and Mustafa, Basil and Kolesnikov, Alexander and Beyer, Lucas},
  booktitle={Proceedings of the IEEE/CVF international conference on computer vision},
  pages={11975--11986},
  year={2023}
}

@article{su2024roformer,
  title={Roformer: Enhanced transformer with rotary position embedding},
  author={Su, Jianlin and Ahmed, Murtadha and Lu, Yu and Pan, Shengfeng and Bo, Wen and Liu, Yunfeng},
  journal={Neurocomputing},
  volume={568},
  pages={127063},
  year={2024},
  publisher={Elsevier}
}

@inproceedings{grosse2009ground,
  title={Ground truth dataset and baseline evaluations for intrinsic image algorithms},
  author={Grosse, Roger and Johnson, Micah K and Adelson, Edward H and Freeman, William T},
  booktitle={2009 IEEE 12th International Conference on Computer Vision},
  pages={2335--2342},
  year={2009},
  organization={IEEE}
}

@article{blackforest2024flux,
    title={Flux.1 Model Family},
    author={BlackForestLabs},
    url={https://blackforestlabs.ai/announcing-black-forest-labs},
    year={2024}
}

@article{hu2022lora,
  title={Lora: Low-rank adaptation of large language models.},
  author={Hu, Edward J and Shen, Yelong and Wallis, Phillip and Allen-Zhu, Zeyuan and Li, Yuanzhi and Wang, Shean and Wang, Lu and Chen, Weizhu and others},
  journal={ICLR},
  volume={1},
  number={2},
  pages={3},
  year={2022}
}

@article{kingma2014adam,
  title={Adam: A method for stochastic optimization},
  author={Kingma, Diederik P},
  journal={arXiv preprint arXiv:1412.6980},
  year={2014}
}

@inproceedings{GSIR,
  title={Gs-ir: 3d gaussian splatting for inverse rendering},
  author={Liang, Zhihao and Zhang, Qi and Feng, Ying and Shan, Ying and Jia, Kui},
  booktitle={Proceedings of the IEEE/CVF Conference on Computer Vision and Pattern Recognition},
  pages={21644--21653},
  year={2024}
}

@article{GIGS,
	year = {2024},
	author = {Chen, Hongze and Lin, Zehong and Zhang, Jun},
	title = {GI-GS: Global Illumination Decomposition on Gaussian Splatting for Inverse Rendering},
	journal = {arXiv preprint arXiv:2410.02619}
}

@inproceedings{yao2022neilf,
  title={Neilf: Neural incident light field for physically-based material estimation},
  author={Yao, Yao and Zhang, Jingyang and Liu, Jingbo and Qu, Yihang and Fang, Tian and McKinnon, David and Tsin, Yanghai and Quan, Long},
  booktitle={European conference on computer vision},
  pages={700--716},
  year={2022},
  organization={Springer}
}

@inproceedings{cache,
  title={Flash cache: Reducing bias in radiance cache based inverse rendering},
  author={Attal, Benjamin and Verbin, Dor and Mildenhall, Ben and Hedman, Peter and Barron, Jonathan T and O’Toole, Matthew and Srinivasan, Pratul P},
  booktitle={European Conference on Computer Vision},
  pages={20--36},
  year={2024},
  organization={Springer}
}

@inproceedings{wang2021learning,
  title={Learning indoor inverse rendering with 3d spatially-varying lighting},
  author={Wang, Zian and Philion, Jonah and Fidler, Sanja and Kautz, Jan},
  booktitle={Proceedings of the IEEE/CVF International Conference on Computer Vision},
  pages={12538--12547},
  year={2021}
}

@inproceedings{zhu2022learning,
  title={Learning-based inverse rendering of complex indoor scenes with differentiable monte carlo raytracing},
  author={Zhu, Jingsen and Luan, Fujun and Huo, Yuchi and Lin, Zihao and Zhong, Zhihua and Xi, Dianbing and Wang, Rui and Bao, Hujun and Zheng, Jiaxiang and Tang, Rui},
  booktitle={SIGGRAPH Asia 2022 Conference Papers},
  pages={1--8},
  year={2022}
}

@inproceedings{srinivasan2021nerv,
  title={Nerv: Neural reflectance and visibility fields for relighting and view synthesis},
  author={Srinivasan, Pratul P and Deng, Boyang and Zhang, Xiuming and Tancik, Matthew and Mildenhall, Ben and Barron, Jonathan T},
  booktitle={Proceedings of the IEEE/CVF conference on computer vision and pattern recognition},
  pages={7495--7504},
  year={2021}
}

@article{zhang2021nerfactor,
  title={Nerfactor: Neural factorization of shape and reflectance under an unknown illumination},
  author={Zhang, Xiuming and Srinivasan, Pratul P and Deng, Boyang and Debevec, Paul and Freeman, William T and Barron, Jonathan T},
  journal={ACM Transactions on Graphics (ToG)},
  volume={40},
  number={6},
  pages={1--18},
  year={2021},
  publisher={ACM New York, NY, USA}
}

@inproceedings{10.1145/566570.566575,
author = {Reinhard, Erik and Stark, Michael and Shirley, Peter and Ferwerda, James},
title = {Photographic tone reproduction for digital images},
year = {2002},
isbn = {1581135211},
publisher = {Association for Computing Machinery},
address = {New York, NY, USA},
url = {https://doi.org/10.1145/566570.566575},
doi = {10.1145/566570.566575},
booktitle = {Proceedings of the 29th Annual Conference on Computer Graphics and Interactive Techniques},
pages = {267–276},
numpages = {10},
location = {San Antonio, Texas},
series = {SIGGRAPH '02}
}

@inproceedings{dit,
  title={Scalable diffusion models with transformers},
  author={Peebles, William and Xie, Saining},
  booktitle={Proceedings of the IEEE/CVF international conference on computer vision},
  pages={4195--4205},
  year={2023}
}

@inproceedings{bolduc2025gaslight,
  title={GaSLight: Gaussian Splats for Spatially-Varying Lighting in HDR},
  author={Bolduc, Christophe and Hold-Geoffroy, Yannick and Lalonde, Jean-Fran{\c{c}}ois},
  booktitle={Proceedings of the IEEE/CVF International Conference on Computer Vision},
  pages={29120--29130},
  year={2025}
}

@InProceedings{Bolduc_2026_CVPR,
    author    = {Bolduc, Christophe and Philip, Julien and Ma, Li and He, Mingming and Debevec, Paul and Lalonde, Jean-Fran\c{c}ois},
    title     = {Lighting in Motion: Spatiotemporal HDR Lighting Estimation},
    booktitle = {Proceedings of the IEEE/CVF Conference on Computer Vision and Pattern Recognition (CVPR)},
    month     = {June},
    year      = {2026},
    pages     = {19696-19705}
}

@article{gera2022casual,
  title={Casual indoor HDR radiance capture from omnidirectional images},
  author={Gera, Pulkit and Dastjerdi, Mohammad Reza Karimi and Renaud, Charles and Narayanan, PJ and Lalonde, Jean-Fran{\c{c}}ois},
  journal={arXiv preprint arXiv:2208.07903},
  year={2022}
}

@inproceedings{Somanath2021HDREM,
  title     = {HDR Environment Map Estimation for Real-Time Augmented Reality},
  author    = {Somanath, Gowri and Kurz, Daniel},
  booktitle = {Proceedings of the IEEE/CVF Conference on Computer Vision and Pattern Recognition (CVPR)},
  year      = {2021},
  pages     = {11293-11301}
}

%
\clearpage
\appendix

\section{Additional Experiments and Visualization Results}
\subsection{Dynamic range comparison on the Laval Indoor dataset and Laval Outdoor dataset}
To further validate our design choice of using only Reinhard tone mapping, we evaluate whether it preserves a sufficient dynamic range in most cases. We compare the illumination estimated by our method and the baselines on the Laval Indoor and Laval Outdoor datasets using three metrics: dynamic range error, Jensen--Shannon (JS) divergence, and Earth Mover's Distance (EMD). Specifically, $\mathrm{DR}_{\mathrm{error}} = \left| \mathrm{DR}_{\mathrm{pred}} - \mathrm{DR}_{\mathrm{GT}} \right|$, $\mathrm{DR} = \log_2 \frac{L{99.9}}{L_{0.1}}$, where $L_{99.9}$ and $L_{0.1}$ denote the 99.9th and 0.1th percentile luminance values, respectively. JS divergence and EMD are computed between the normalized luminance histograms of the predicted and ground-truth illumination.

Table~\ref{tab:DR} reports the quantitative results. Our method consistently outperforms other generative lighting-estimation approaches, achieving the best overall performance across both datasets. These results confirm that our design choice is sufficient for preserving the dynamic range in most scenarios.

\begin{table}[h]
    \centering
    \caption{Quantitative comparison for dynamic range on the Laval Indoor and Laval Outdoor datasets. Lower is better for all metrics.}
    \label{tab:DR}
    \begin{tabular}{llccc}
        \toprule
        Scene & Method & DR-Error $\downarrow$ & JS $\downarrow$ & EMD $\downarrow$ \\
        \midrule
        \multirow{4}{*}{Laval Indoor}
        & StyleLight     & 5.541 & 0.097 & 0.718 \\
        & DiffusionLight & 3.538 & 0.094 & 0.477 \\
        & LuxDiT         & \underline{2.960} & \underline{0.082} & \underline{0.471} \\
        & Ours           & \textbf{2.626} & \textbf{0.059} & \textbf{0.379} \\
        \midrule
        \multirow{4}{*}{Laval Outdoor}
        & StyleLight     & 9.996 & 0.106 & 0.663 \\
        & DiffusionLight & \textbf{1.688} & \underline{0.102} & \underline{0.460} \\
        & LuxDiT         & 2.235 & 0.122 & 0.547 \\
        & Ours           & \underline{2.103} & \textbf{0.100} & \textbf{0.410} \\
        \bottomrule
    \end{tabular}
\end{table}

To further compare the dynamic range of the predicted illumination, we augment the original three-sphere protocol by adding a diffuse plane beneath the spheres, with an albedo of \([0.8, 0.8, 0.8]\) and a roughness of \(0.5\). We render the scene using a camera elevation of \(45^\circ\) and compute the corresponding rendering metrics in this shadow-aware setting. Table~\ref{tab:three_spheres_shadow} reports the quantitative results. Our method continues to outperform other generative lighting-estimation approaches, further supporting our claim that Reinhard tone-mapping preserves sufficient dynamic range in most scenarios.

\begin{table*}[h]
\centering
\large
\caption{Quantitative results on the Laval Indoor and Laval Outdoor datasets for the 3-Sphere-Shadow setting with shadow. Underlined entries indicate the second-best results.}
\vspace{-9pt}
\label{tab:three_spheres_shadow}
\setlength{\tabcolsep}{6.8pt}
\renewcommand{\arraystretch}{0.8}
\begin{tabular}{
    l
    l
    c
    c
    c
    c
    c
    c
    c
    c
    c
}
\toprule
\multirow{2}{*}{\textbf{Dataset}} & \multirow{2}{*}{\textbf{Method}} & \multicolumn{3}{c}{\textbf{Scale-invariant RMSE} $\downarrow$} & \multicolumn{3}{c}{\textbf{Angular Error} $\downarrow$}  & \multicolumn{3}{c}{\textbf{Normalized RMSE} $\downarrow$}                                                            \\
 & & Diffuse           & Matte           & Mirror           & Diffuse           & Matte           & Mirror   & Diffuse           & Matte           & Mirror      \\ 
 [-2pt]
\midrule
\multirow{4}{*}{Laval Indoor}
& StyleLight
& 0.031 & 0.049 & \underline{0.084}
& 3.861 & 3.908 & 3.997
& 0.166 & 0.259 & 0.328 \\
& DiffusionLight
& 0.030 & 0.050 & 0.094
& 3.193 & 3.236 & 3.318
& 0.160 & \textbf{0.211} & \textbf{0.233} \\
& LuxDiT
& \underline{0.026} & \underline{0.043} & 0.086
& \underline{3.014} & \underline{3.066} & \underline{3.137}
& \underline{0.140} & 0.227 & 0.303 \\
& Ours
& \textbf{0.025} & \textbf{0.041} & \textbf{0.083}
& \textbf{2.957} & \textbf{3.011} & \textbf{3.101}
& \textbf{0.126} & \underline{0.212} & \underline{0.257} \\
\hline
\\[-9pt]

\multirow{4}{*}{Laval Outdoor}
& StyleLight
& 0.099 & 0.122 & 0.171
& 5.922 & 5.983 & 6.140
& 0.269 & 0.253 & 0.239 \\
& DiffusionLight
& \underline{0.079} & \underline{0.105} & \underline{0.162}
& 4.386 & 4.455 & 4.574
& \underline{0.207} & \underline{0.217} & \underline{0.229} \\
& LuxDiT
& 0.088 & 0.127 & 0.208
& \underline{4.362} & \underline{4.421} & \underline{4.532}
& 0.279 & 0.264 & 0.343 \\
& Ours
& \textbf{0.048} & \textbf{0.075} & \textbf{0.136}
& \textbf{2.606} & \textbf{2.656} & \textbf{2.781}
& \textbf{0.145} & \textbf{0.177} & \textbf{0.205} \\
[-3pt]
\bottomrule

\end{tabular}
\vspace{-7pt}
\end{table*}

\subsection{Array of Spheres}
We adopt the array-of-spheres protocol used in prior works~\cite{dastjerdi2023everlight, Phongthawee_2024_CVPR} to further evaluate performance on the Laval Indoor dataset. Following these works, we use the same 2,240 perspective crops extracted from 224 HDRI maps. For each crop, we estimate the corresponding HDRI map and use it to render an array of diffuse spheres placed on a ground plane.

\begin{figure*}[h]
    \centering
    \includegraphics[width=0.99\linewidth]{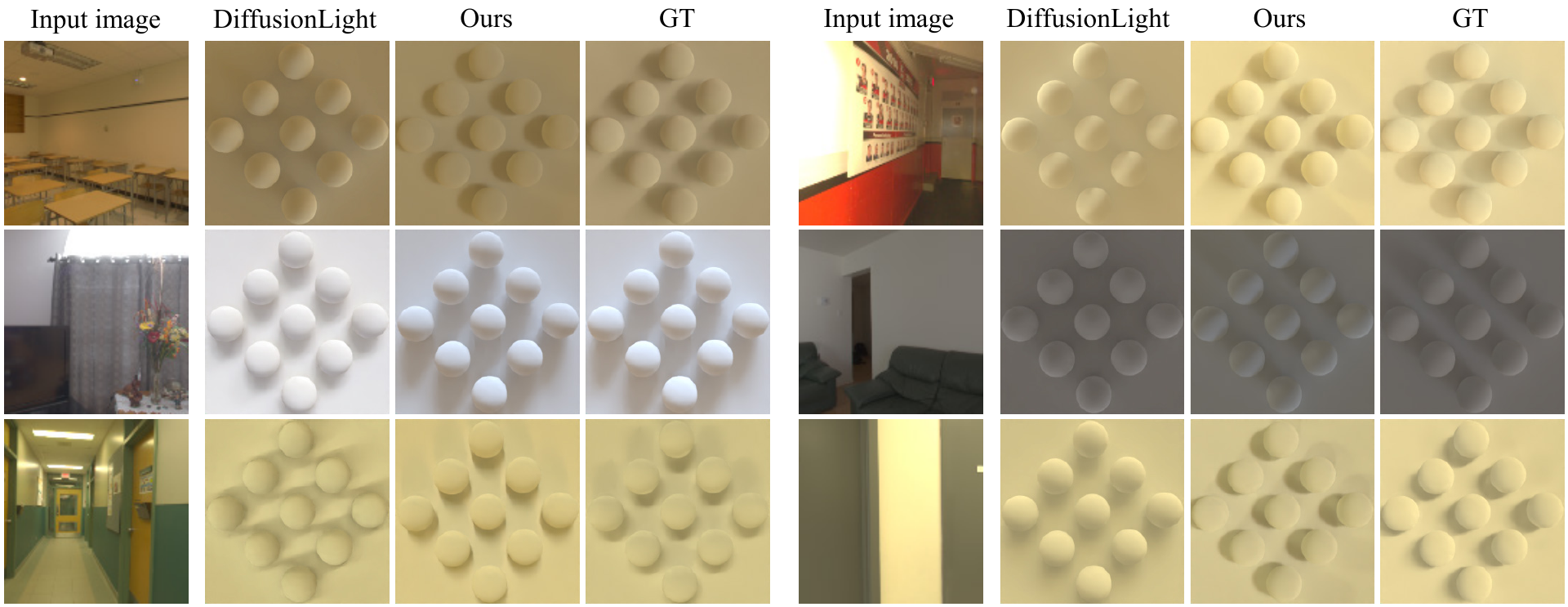} 
    \vspace{-10pt}
    \caption{Qualitative results under the array of spheres protocol on the Laval Indoor dataset.}
    \vspace{-5pt}
    \label{fig:array of shperes}
\end{figure*}

Table \ref{tab:array of shpere} reports the quantitative results, where our method outperforms other generative lighting estimation approaches. Fig. \ref{fig:array of shperes} shows qualitative comparisons, demonstrating that \name{} recovers more accurate light-source locations than DiffusionLight. 
\begin{table}[h]
\centering
\large
\caption{Quantitative results under the array of spheres protocol on the Laval Indoor dataset.}
\vspace{-10pt}
\label{tab:array of shpere}
\setlength{\tabcolsep}{12pt}
\begin{tabular}{lcc}
\toprule
\textbf{Method} & \textbf{si-RMSE} $\downarrow$ & \textbf{Angular-Error} $\downarrow$ \\
\midrule
EverLight       & 0.091 & 6.360 \\
StyleLight      & 0.123 & 7.090 \\
DiffusionLight  & 0.090 & 5.250 \\
LuxDiT          & \underline{0.089} & \underline{4.900} \\
Ours            & \textbf{0.087} & \textbf{4.474} \\
\bottomrule
\end{tabular}
\end{table}

\subsection{Spatially-varying illumination Estimation Comparison on the PanoHDR-NeRF Dataset}

We further evaluate our method on the PanoHDR-NeRF dataset, including two additional baselines: PanoHDR-NeRF and GasLight. Both approaches follow a two-stage formulation. Specifically, they first employ a network to reconstruct HDR images from the input LDR observations, and subsequently optimize a radiance field in linear HDR space using the predicted HDR images. HDRI maps are then rendered from the optimized radiance field.

As shown in Table \ref{tab:panohdr}, our method achieves better quantitative performance than both baselines, further validating its effectiveness and generalization to real-world scenes. In addition, both PanoHDR-NeRF and GasLight rely on dense multi-view inputs for radiance field reconstruction, whereas our method naturally extends to sparse-view settings, including the single-view case, thereby offering a more practical solution.

\begin{table}[h]
\centering
\large
\caption{Quantitative comparison for spatially-varying lighting estimation on PanoHDR-NeRF Dataset.}
\vspace{-10pt}
\label{tab:panohdr}
\setlength{\tabcolsep}{12pt}
\begin{tabular}{lcc}
\toprule
\multirow{2}{*}{\textbf{Method}} & \textbf{Chess room} & \textbf{Small class} \\
 & \textbf{PU21-PSNR} $\downarrow$ & \textbf{PU21-PSNR} $\downarrow$ \\
\midrule
PanoHDR-NeRF & 27.61 & 28.57 \\
GasLight & 30.29 & 30.77 \\
Ours & \textbf{31.28} & \textbf{34.70} \\
\bottomrule
\end{tabular}
\end{table}

\subsection{Additional Visualization Results of Spatially-Varying Illumination Estimation on Real-World Captures}
\begin{figure*}[h]
    \centering
    \includegraphics[width=0.99\linewidth]{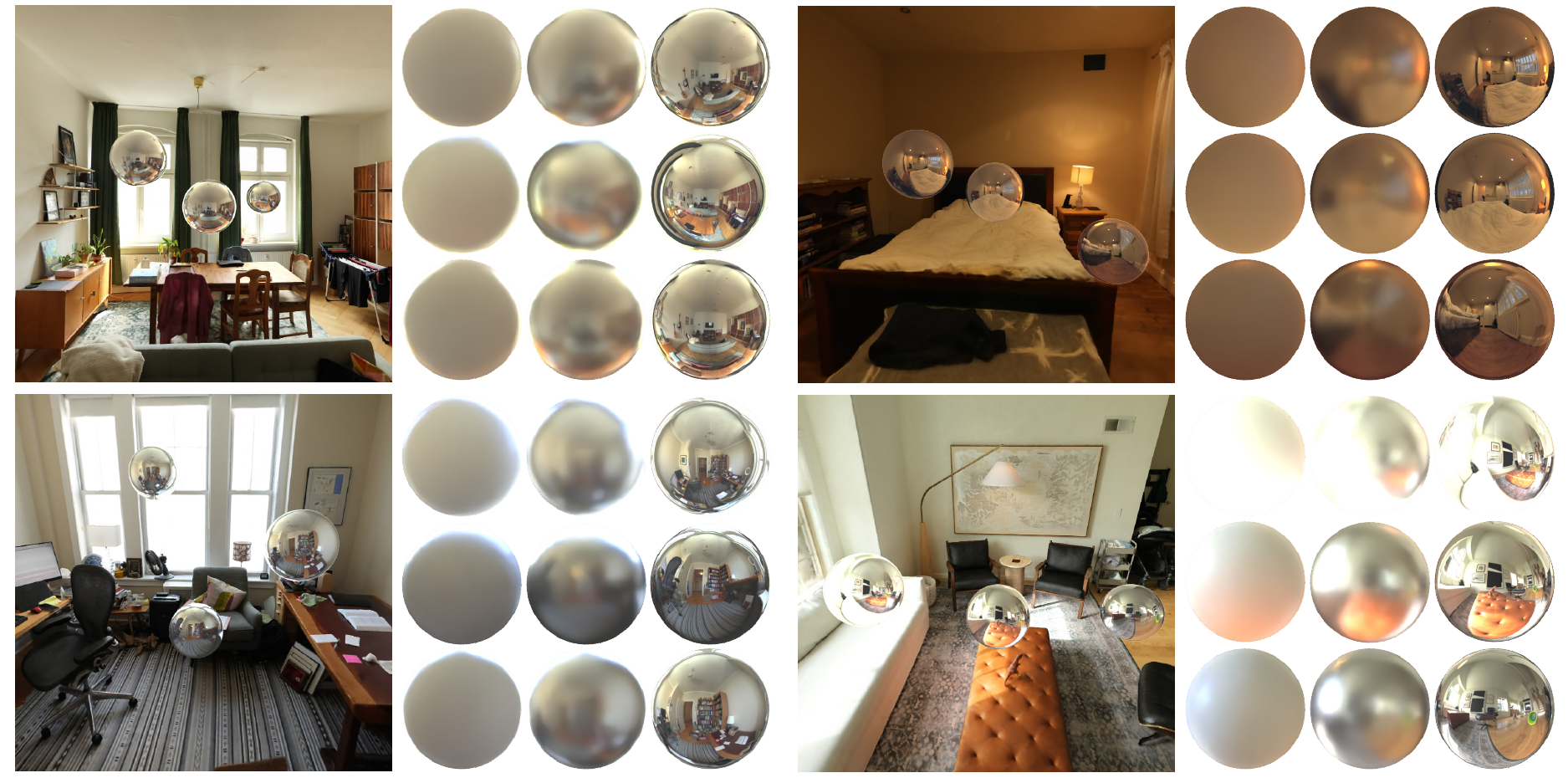} 
    \vspace{-10pt}
    \caption{Visualization of the estimated spatially-varying illumination on in-the-wild scenes. The inserted spheres on the left correspond one-to-one to the rendered results on the right: the spheres are ordered from left to right, matching the results arranged from top to bottom.}
    \vspace{-5pt}
    \label{fig:real1}
\end{figure*}

\begin{figure*}[h]
    \centering
    \includegraphics[width=0.99\linewidth]{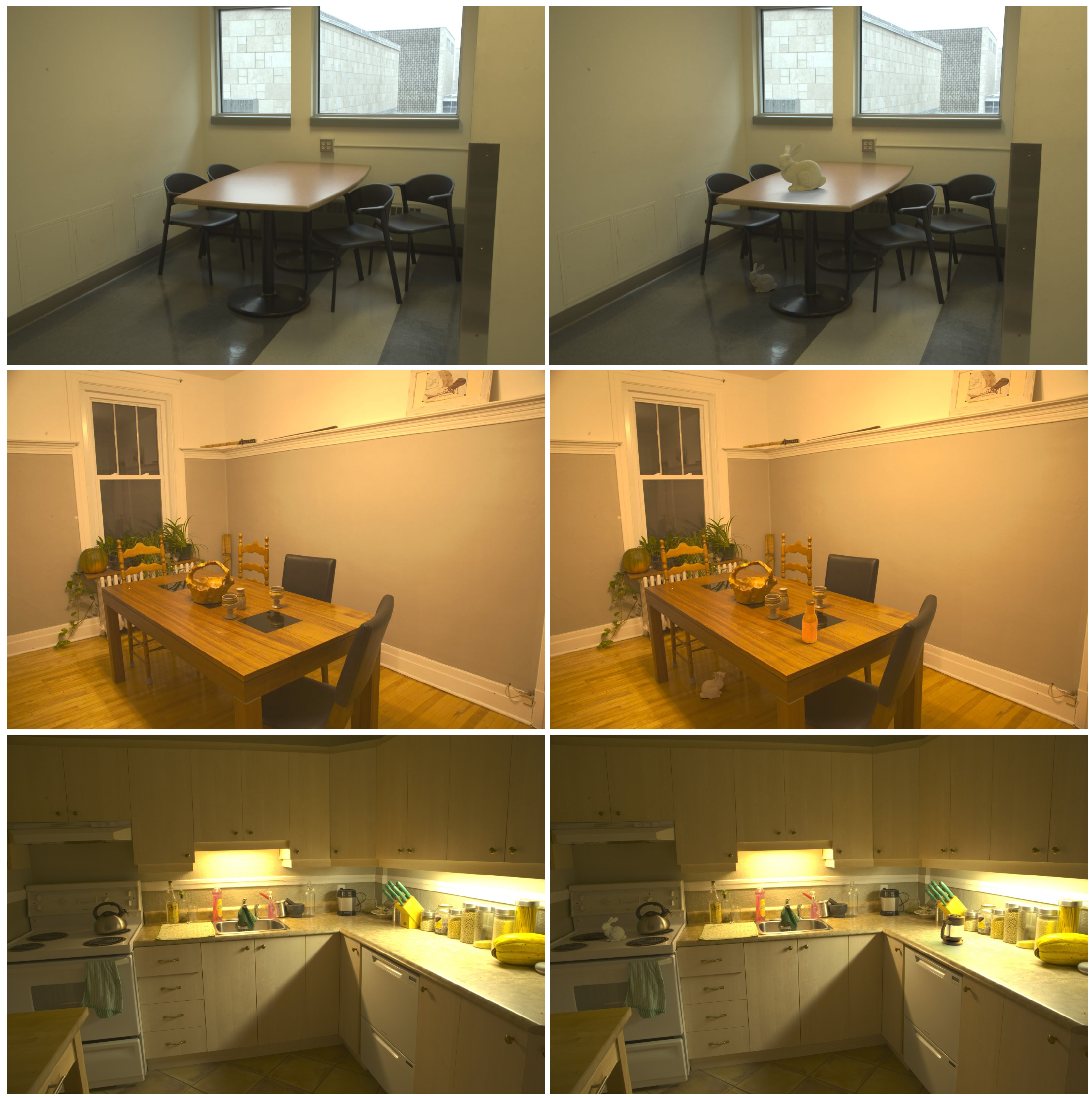} 
    \vspace{-10pt}
    \caption{Visualization of the object insertion results on the Laval Indoor Spatially Varying dataset.}
    \vspace{-5pt}
    \label{fig:real2}
\end{figure*}

We present spatially-varying illumination estimation results on in-the-wild, real-world captures in Figs. \ref{fig:real1} and Figs. \ref{fig:real2}. As shown in Fig.~\ref{fig:real1}, \name{} produces spatially varying illumination that is both consistent and visually faithful, with fine-grained details. For example, in the bottom-right case, as the sphere moves farther away from the window, the predicted illumination gradually weakens, while the spatial distribution of the illumination remains well aligned with the scene. This result demonstrates strong performance and robust generalization to unconstrained real-world scenes. In addition, the object insertion results in Fig.~\ref{fig:real2} further demonstrate the accuracy of our predictions, as the inserted objects are naturally and seamlessly integrated into the scene.

\subsection{HDRI maps Synthesis Results}
\begin{figure*}[h]
    \centering
    \includegraphics[width=0.99\linewidth]{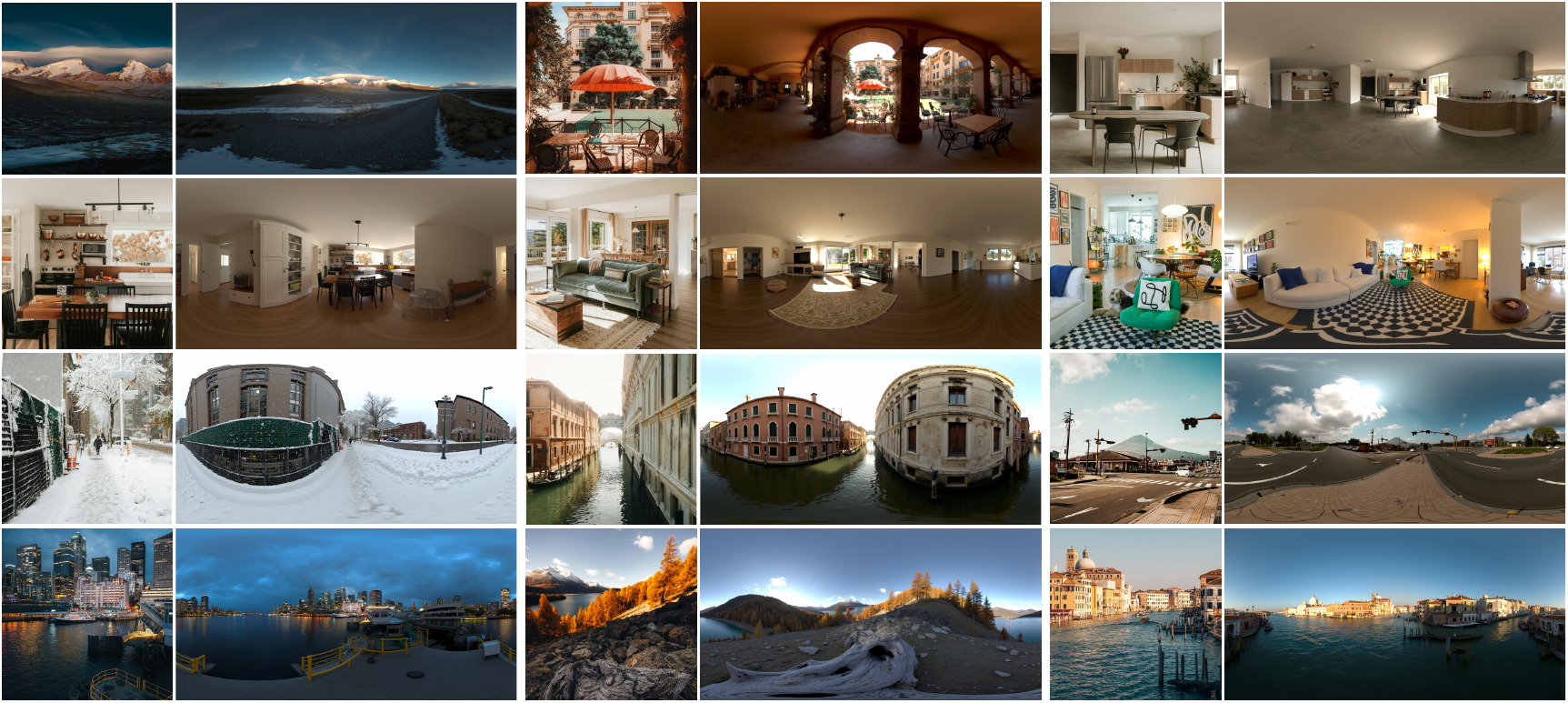} 
    \vspace{-10pt}
    \caption{Image conditioned panorama generation results on real-world captures.}
    \vspace{-5pt}
    \label{fig:inthewild}
\end{figure*}

Fig. \ref{fig:inthewild} demonstrates image-conditioned HDRI map synthesis for real-world photographs. The generated panoramas achieve photorealistic quality across diverse indoor and outdoor scenarios with  coherent scene layouts and illumination distributions, highlighting the strong generalizability of our approach. 

\section{Dataset Details}

\subsection{Dataset Curation}
\paragraph{Synthetic Dataset} We conduct a synthetic dataset that contains around 700 rooms of five types, including bedroom, bathroom, dining room, kitchen, and living room. Moreover, we render the multi-view images and light probes using Blender Cycles \footnote{https://www.blender.org/} renderer with 1024 samples per pixel and OptiX denoiser. It takes approximately six days to generate the scenes and another four days to render the dataset. The overall dataset contains around 8,500 RGB images and 15,000 HDRI maps with ground-truth camera parameters and geometry. 

We further provide details on the light probe placement strategy. For each scene, we first extract the meshes corresponding to the walls, floor, and ceiling, and merge them into a closed watertight mesh that serves as the bounding box. Candidate light probe locations are then uniformly sampled within this bounding box. For each sampled location, we cast 128 rays uniformly distributed over the sphere and measure their travel distances. If the minimum ray distance falls below a predefined threshold, the sampled location is regarded as being too close to nearby occluders and is therefore rejected. We repeat this process until the target number of valid probe locations is obtained. In addition, we uniformly sample extra probe locations within the camera frustum and apply the same occlusion criterion to filter out invalid samples.
\paragraph{Real-world Dataset} We collect real-world HDRI maps from the following four public databases:

\begin{itemize}
\item \emph{Poly Haven \footnote{https://polyhaven.com/hdris}}, which contains 944 HDRI maps of diverse indoor and outdoor scenes.
\item \emph{Ambient CG \footnote{https://ambientcg.com/}}, which contains 375 HDRI maps of diverse indoor and outdoor scenes.
\item \emph{HDR Maps \footnote{https://hdrmaps.com/}}, which contains 186 HDRI maps of indoor and outdoor scenes.
\item \emph{HDRI Skies \footnote{https://hdri-skies.com/}}, which contains 22 HDRI maps of outdoor scenes.
\end{itemize}

\subsection{Implementation Details}
For all baselines, we use the official publicly released codebases and pretrained weights to obtain the reported results. For StyleLight, we further fine-tune the pretrained model on the Laval Outdoor dataset for 25,000 iterations to make it applicable to outdoor scenes.

For 4DLighting, the official codebase provides a framework for optimizing an MLP light field from a single-view input using SDS. Specifically, under the camera coordinate system of the input image, the method samples nine different positions within the camera frustum at each iteration, uses the diffusion model to predict the corresponding light probes, and optimizes the MLP light field based on these predictions.

To extend this framework to the multi-view setting, we first estimate the camera poses of the input multi-view images and normalize them to the coordinate system of the first camera. We then randomly select one input view and similarly sample nine positions within its camera frustum, from which the diffusion model predicts the corresponding light probes. The MLP light field is then optimized according to the 3D locations of these nine probes expressed in the coordinate system of the first camera. In addition, while the original 4DLighting represents light probe locations using pixel coordinates and inverse depth, we instead use the 3D coordinates of each light probe in the multi-view setting to preserve correct spatial and projection relationships.

\end{document}